%% file: optimal_transport_compression.tex
\documentclass{article}
\usepackage{amsmath,amssymb,stmaryrd}
\usepackage{amsthm}
\usepackage{authblk}
\usepackage{natbib}
    \usepackage[english]{babel}
    \usepackage[utf8]{inputenc}
\usepackage{color}

\usepackage{mathtools}
\DeclarePairedDelimiter\ceil{\lceil}{\rceil}
\DeclarePairedDelimiter\abs{\lvert}{\rvert}

\usepackage{algorithm2e}

\usepackage{caption}
\usepackage{subcaption}
\usepackage{algorithmic}
\newtheorem{theorem}{Theorem}

\usepackage{subcaption}
\usepackage[title]{appendix}

\newif\ifsolns
\solnsfalse

\ifsolns
\newcommand{\soln}[1]{\newline \noindent {\bfseries Solution:} {\itshape #1}}
\else
\newcommand{\soln}[1]{}
\fi
\usepackage[margin=2.75cm]{geometry}

\usepackage{fancybox}
\begin{document}
\title{A streaming feature-based compression method for
  data from instrumented infrastructure
}
\author[1,2]{Alastair Gregory}
\author[1,2]{Din-Houn Lau}
\author[3]{Alex Tessier}
\author[3]{Pan Zhang}
\affil[1]{Lloyd's Register Foundation's Programme for Data-Centric Engineering,
Alan Turing Institute}
\affil[2]{
Department of Mathematics, Imperial College London, UK % \footnote{Department of Mathematics, Imperial College London, Exhibition Road, South Kensington, SW7 2AZ.}
}
\affil[3]{Autodesk Research}

\maketitle

\begin{abstract}

An increasing amount of civil engineering applications are utilising data acquired from infrastructure instrumented with sensing devices. This data has an important role in monitoring the response of these structures to excitation, and evaluating structural health. In this paper we seek to monitor pedestrian-events (such as a person walking) on a footbridge using strain and acceleration data. The rate of this data acquisition and the number of sensing devices make the storage and analysis of this data a computational challenge.  We introduce a streaming method to compress the sensor data, whilst preserving key patterns and features (unique to different sensor types) corresponding to pedestrian-events. Numerical demonstrations of the methodology on data obtained from strain sensors and accelerometers on the pedestrian footbridge are provided to show the trade-off between compression and accuracy during and in-between periods of pedestrian-events.

\end{abstract}

\section{Introduction}\label{sec:introduction}

\input{intro}

\section{Strain and accelerometer data for instrumented infrastructure}\label{sec:data}

\input{motivation}

\section{Relevance scores for features in time-series}\label{sec:relevancefunctions}

\input{relevance}

\section{Segmentation and compression}\label{sec:time-seri-segm}
\input{segmentation}

\section{Streaming time-series segmentation}\label{sec:streamingdata}

\input{streaming_segmentation}

\section{Numerical demonstrations}\label{sec:numer-demonstr}

\input{numerics}

\section{Conclusion and Discussion}\label{sec:concl-disc}
\input{conclusion}

%\section{Compression and time-series segmentation}

%Consider the time-series $\big\{(x_{i},y_i)\big\}_{i=1}^{n}$ where $x_i$ denote the time stamps that the data $y_i$ is available at. 

\bibliography{refs}

\begin{appendix}

\section{Linear programming algorithm}
\label{sec:algolp}

The algorithm for linear programming is given in Algorithm \ref{alg:general}.

\begin{algorithm}[tbh]
        \caption{Linear programming for constructing the coupling matrix $\eta$}\label{alg:general}
        \begin{algorithmic}[1]
            \REQUIRE $w_1,w_2,\ldots,w_n$ and $\tilde{w}_1,\tilde{w}_2,\ldots,\tilde{w}_{n'}$;

            \STATE Set $i=n$ and $j=n'$;

            \WHILE{$i \times j > 0$}
			\IF{$w_i \leq\tilde{w}_j$}
			\STATE $\eta_{i,j}=w_i$;
			\STATE $\tilde{w}_j=\tilde{w}_j-w_i$;
			\STATE $i = i-1$;
			\ENDIF
			\IF{$\tilde{w}_j < w_i$}
			\STATE $\eta_{i,j}=\tilde{w}_{j}$;
			\STATE $w_i=w_i-\tilde{w}_j$;
			\STATE $j = j-1$;
			\ENDIF
			\ENDWHILE
\RETURN $\eta$
\end{algorithmic}
\end{algorithm}

\section{Proof of the error in relevance of reconstructions}

\label{sec:showingreconstruction}

This section explains the derivation of the error bound in the relevance scores of the compressed reconstruction with respect to that of the original time-series, given in (\ref{equation:reconstructionerrorbound}). We start by assuming smoothness of the relevance score $\phi_i$,
$$
\tilde{\phi}_i \in \left[\min_{x_k \in [\tilde{x}_j,\tilde{x}_{j+1}]}\phi_k,\max_{x_k \in [\tilde{x}_j,\tilde{x}_{j+1}]}\phi_k\right],
$$
where $x_i \in [\tilde{x}_j, \tilde{x}_{j+1}]$. Let $Z=\sum^{n}_{i = 1}\phi_i$. We will now assume that the piecewise linear reconstruction in (\ref{equation:piecewiselinear}) is used, and a relevance score $\phi_i$ that just depends on $y_i$, for $i=1,\ldots,n$, is used. We then investigate two cases: \textbf{(a)} $\max_{x_k \in [\tilde{x}_j,\tilde{x}_{j+1}]}\phi_k < Z/n'$, \textbf{(b)} $\max_{x_k \in [\tilde{x}_j,\tilde{x}_{j+1}]}\phi_k \geq Z/n'$. For case \textbf{(a)}, it is clear that at any point $x_i \in [\tilde{x}_j,\tilde{x}_{j+1}]$, the error of the relevance score for the reconstructed time-series is,
$$
\abs*{\tilde{\phi}_i-\phi_i} \leq \max_{x_k \in [\tilde{x}_j,\tilde{x}_{j+1}]}\phi_k < \frac{Z}{n'},
$$
assuming $\phi \geq 0$. Now for case \textbf{(b)}, we know that the last point in the interval $[\tilde{x}_j,\tilde{x}_{j+1}]$ will be the value of $x_l$, where $\phi_l=\max_{x_k \in [\tilde{x}_j,\tilde{x}_{j+1}]}\phi_k$. Since a piecewise linear approximation is assumed, we have $S\left(\tilde{x}_{j+1};\big\{\tilde{x}_j\big\}_{j=1}^{n'}\right)=y_{j+1}$, and therefore
$$
\abs*{\tilde{\phi}_{l}- \phi_{l}} = 0.
$$
We also note that of course $\max_{x_k \in [\tilde{x}_j,\tilde{x}_{j+1}-1]}\phi_k< \frac{Z}{n'}$. Then at any time-stamp $x_i \in [\tilde{x}_j,\tilde{x}_{j+1}]$, the error of the relevance score for the reconstructed time-series is,
$$
\abs*{\tilde{\phi}_i-\phi_i} \leq
\max_{x_k \in [\tilde{x}_j,\tilde{x}_{j+1}-1]}\phi_k < \frac{Z}{n'}.
$$
Therefore for either case, and for the assumptions placed on the reconstruction, we have that
$$
\abs*{\tilde{\phi}_i-\phi_i} < \frac{Z}{n'},
$$
for all $i =1,\ldots,n$.

\qed

\section{Streaming approximation to segmentation points}

\label{sec:appendixstreamingdata}

The approximation to the segmentation points outlined in Sec. \ref{sec:streamingdata} is explained in more detail here. The construction of the approximation is based on storing a synopsis of the data points in the time-series, and is inspired by the work in \cite{Greenwald}. The synopsis is a set $S$ formed of the triples $(x_i^S,\phi^S_i,\xi_i)$, for $i=1,\ldots,L$, where the values $x^S_i \in \big\{x_{l}\big\}_{l=1}^{n}$, for $i=1,\ldots,L$, are a succinct collection of time-stamps within the time-series. They are such that $x^S_i\leq x^S_{i+1}$, with $x^S_1=x_1$ and $x^S_L=x_{n}$. The values $\big\{\phi^S_{i}\big\}_{i=1}^{L}$ represent the sum of $\phi_l$ over all the time-stamps $x_l \in (x^S_{i-1},x^S_i]$. Finally the values $\big\{\xi_i\big\}_{i=1}^{L}$ represent the sum of the products of $x_l$ and $\phi_l$,  over all the time-stamps $x_l \in (x^S_{i-1},x^S_i]$. The approximation is more efficient than re-computing the segmentation points via Algorithm \ref{alg:general} since the approximation operates on only the data points stored in this synopsis, and given that $L \ll n$. The approximation starts by initializing the values $n$, $n'$, $Z$ and $\Delta Z$ in step (1) of the outline in Sec \ref{sec:streamingdata}. These triples are maintained over time to generate the approximations $\big\{\tilde{x}_j^S\big\}_{j=1}^{n'}$, to the segmentation points $\big\{x_j\big\}_{i=1}^{n'}$, using the Algorithms \ref{alg:inserting}, \ref{alg:combining} and \ref{alg:update} below. Then, an approximation to the segmentation points $\tilde{x}_j$, for $j=1,\ldots,n'$, can be queried at any time via Algorithm \ref{alg:query}.
\iffalse
; these approximations satisfy:
\begin{equation}
\abs*{\tilde{x}_j^S-\tilde{x}_j}/\tilde{x}^S_j \leq 4 \epsilon n',
\label{equation:boundsegpoints}
\end{equation}
for a pre-defined $\epsilon$. This bound is proved in Appendix \ref{sec:proof}. It makes sense to allow $\epsilon=\alpha/n'$, with $\alpha=\mathcal{O}(1)$, and therefore this approximation has constant-bounded relative error of $4\alpha$ for any number of segmentation points. Therefore this is an approximation that maintains a bounded \textit{relative} error. More importantly, the approximation guarantees a bound on the sum of relevance scores for the points inbetween the approximated segmentation point and the actual segmentation point, i.e.
\begin{equation}
\abs*{\sum^{\tilde{x}_j^S}_{i=1}\phi_i-\sum^{\tilde{x}_j}_{i=1}\phi_i} \leq \epsilon\left(\sum^{n}_{i=1}\phi_i\right).
\end{equation}
\fi
%The algorithm to compute this approximation assumes that the first $n$ points $\big\{(x_{i},y_{i})\big\}_{i=1}^{n}$ are known in advance, and that the values $x_i$, for $i=1,2,3,\ldots$, are increasing values.
The triples are maintained as follows. Every time a new element $y_{n+1}$ is added to the time-series at the time-stamp $x_{n+1}$, Algorithm \ref{alg:update} is implemented to update the synopsis. This routine uses Algorithm \ref{alg:inserting} and Algorithm \ref{alg:combining}; the latter algorithm allows the synopsis to be cut down in size in order to make the approximation efficient.

\begin{algorithm}[tbh]
        \caption{Insertion into the synopsis}\label{alg:inserting}
        \begin{algorithmic}[1]
            \REQUIRE $\phi_i$ when the point $y_i$ becomes available in the time-series (or $\boldsymbol y_{(i-\gamma):(i+\gamma)}$ using a buffer, if required for a query-driven relevance score) at the time-stamp $x_i$; $S$

			\STATE Define the triple $t^*=(x_i,\phi_i,x_i\phi_i)$

            \STATE Assuming $x_i>x_j$, $\forall j \in [1, i-1]$, insert $t^*$ on to the end of $S$.

\RETURN $S$
\end{algorithmic}
\end{algorithm}

% algorithm for combining

\begin{algorithm}[tbh]
        \caption{Pruning the synopsis}\label{alg:combining}
        \begin{algorithmic}[1]
            \REQUIRE $\epsilon$; $S$;

			\STATE $j=L-1$;
			\IF{$L > 3$}
            \WHILE{$j \geq 3$}
			\STATE If $\phi^S_j \leq \epsilon \sum^{L}_{l=1}\phi^S_l$ then find $i = \text{arg} \min_{k \in [2, j]}\left(\sum^j_{l=k}\phi^S_l \leq \epsilon \sum^L_{l=1} \phi^S_l\right)$;
			\STATE Combine the triples $(x^S_i,\phi^S_i,\xi_i),\ldots,(x^S_j,\phi^S_j,\xi_j)$ into the new triple $(x^S_j,\sum^j_{l=i}\phi^S_l,\sum^{j}_{l=i}\xi_l)$ in $S$;
			\STATE Set $j=i-1$.
			\ENDWHILE
			\ENDIF
\RETURN $S$
\end{algorithmic}
\end{algorithm}

% algorithm for whole stream

\begin{algorithm}[tbh]
        \caption{Updating the synopsis when a new data point arrives in time-series}\label{alg:update}
        \begin{algorithmic}[1]
            \REQUIRE $\phi_i$ when the point $y_i$ becomes available in the time-series (or $\boldsymbol y_{(i-\gamma):(i+\gamma)}$ using a buffer, if required for a query-driven relevance score) at the time-stamp $x_i$; $\alpha$; $Z$; $\Delta Z$; $n'$; $S$;

			\IF{$\Delta Z \geq Z / n'$}
			\STATE Set $Z=Z+\Delta Z$;
			\STATE Set $\Delta Z = 0$;
			\STATE Set $n'=n'+1$;
			\STATE Set $\epsilon=\alpha/n'$;
			\STATE Run Algorithm \ref{alg:combining} with $\epsilon$; $S$
			\ENDIF
			\IF{$\Delta Z < Z / n'$}
			\STATE Set $\Delta Z = \Delta Z + \phi_i$;
			\ENDIF

			\STATE Run Algorithm \ref{alg:inserting} with $S$.

\RETURN $Z$; $\Delta Z$; $n'$; $S$
\end{algorithmic}
\end{algorithm}

% algorithm for querying

\begin{algorithm}[tbh]
        \caption{Querying the synopsis for approximations to the segmentation points $\tilde{x}_1^S,\tilde{x}_2^S,\ldots,\tilde{x}_{n'}^S$}\label{alg:query}
        \begin{algorithmic}[1]
            \REQUIRE $n'$; $S$;
			\STATE Compute $\Phi^S_i=\sum^i_{l=1}\phi^S_l$, for $i=1,\ldots,L$, and set $\Phi^S_0=0$;
			\FOR{$j=1,\ldots,n'$}
			\STATE Find $i_l$ satisfying $\Phi^S_{i_l-1} < (j-1)\big(\sum^{L}_{l=1}\phi^S_l\big)/n' \leq \Phi^S_{i_l}$;
			\STATE Find $i_u$ satisfying $\Phi^S_{i_u-1} < j\big(\sum^{L}_{l=1}\phi^S_l\big)/n' \leq \Phi^S_{i_u}$;
			\STATE Compute $e_1=x^S_{i_l}\Big(\sum^{i_l}_{i=1}\phi^S_i-(j-1)\big(\sum^{L}_{l=1}\phi^S_l\big)/n'\Big)$;
			\STATE Compute $e_2=x^S_{i_u}\Big(\sum^{i_u}_{i=1}\phi^S_i-j\big(\sum^{L}_{l=1}\phi^S_l\big)/n'\Big)$;
			\STATE Define the approximation $\tilde{x}^{S}_{j}=\Big(\big(\sum^{i_u}_{l=i_l}\xi_l\big)+e_1-e_2\Big)/\big(Z/n'\big)$;
			\ENDFOR

\RETURN $\tilde{x}_1^S,\tilde{x}_2^S,\ldots,\tilde{x}_{n'}^S$
\end{algorithmic}
\end{algorithm}

\section{Proof of streaming approximation error}

\label{sec:proof}

This section provides a proof of the bound given in (\ref{equation:errorboundsegpoints}) for the error of the approximation to the segmentation points $\tilde{x}_j$, for $j=1,\ldots,n'$. First, let $Z=\sum^{n}_{i=1}\phi_i$ and recall that the indices $k_{j-1}$ and $k_j$ are the smallest and largest non-zero elements in $\eta_{1:n,j}$ respectively. Also recall that $k_{j-1}=\text{arg}\min_l \big\{\sum^l_{i=1}\phi_i \geq (j-1)Z/n'\big\}$ and $k_j=\text{arg}\min_l \big\{\sum^l_{i=1}\phi_i \geq jZ/n'\big\}$. Assume that the approximations in Algorithm \ref{alg:query} to the indices $k_{j}$ and $k_{j-1}$ (corresponding to the $i_{u}$'th and $i_l$'th triple in $S$ respectively) are given by $\hat{k}_j$ and $\hat{k}_{j-1}$. Note that due to the way that $S$ is constructed and maintained, we have that if $x^S_{i_u} \neq x_{k_{j}}$ we must have $\phi^S_{i_u} \leq \epsilon Z$, and that if $x^S_{i_l} \neq x_{k_{j-1}}$ we must have $\phi^S_{i_l} \leq \epsilon Z$. Also recall that
$$
\abs*{\sum^{k_{j-1}}_{l=1}\phi_l - \sum^{\hat{k}_{j-1}}_{l=1}\phi_l} \leq \epsilon \left(\sum^{n}_{l=1}\phi_l\right), \quad \abs*{\sum^{k_{j}}_{l=1}\phi_l - \sum^{\hat{k}_{j}}_{l=1}\phi_l} \leq \epsilon \left(\sum^{n}_{l=1}\phi_l\right),
$$
from (\ref{equation:gcondition}). The error of the streaming approximation to the segmentation points $\tilde{x}_j$, for $j=1,\ldots,n'$, can be expressed as,
\begin{equation}
\abs*{\tilde{x}_j-\tilde{x}_j^S} = \abs*{A-\hat{A}},
\label{equation:totalerror}
\end{equation}
where
\begin{equation*}
\begin{split}
A &= \sum^{n}_{i=1}\eta_{i,j}n'x_{i}=\frac{n'}{Z}\sum^{n}_{i=1}Z\eta_{i,j}x_{i}\\
\quad &=\frac{n'}{Z}\left(\left(\sum^{k_j}_{i=k_{j-1}}\phi_ix_i\right)+\left(x_{k_{j-1}}\left(\sum^{k_{j-1}}_{i=1}\phi_i-\frac{(j-1)Z}{n'}\right)\right)-\left(x_{k_{j}}\left(\sum^{k_{j}}_{i=1}\phi_i-\frac{jZ}{n'}\right)\right)\right),
\end{split}
\end{equation*}
and
$$
\hat{A} = \frac{n'}{Z}\left(\left(\sum^{i_u}_{i=i_{l}}\xi_i\right)+\left(x^S_{i_l}\left(\sum^{i_l}_{i=1}\phi^S_i-\frac{(j-1)Z}{n'}\right)\right)-\left(x^S_{i_u}\left(\sum^{i_u}_{i=1}\phi^S_i-\frac{jZ}{n'}\right)\right)\right).
$$
Now, let $\hat{A}_1=\sum^{i_u}_{i=i_{l}}\xi_i$ and $A_1=\sum^{k_j}_{i=k_{j-1}}\phi_ix_i$. Then let $\hat{A}_2=x^S_{i_l}\left(\sum^{i_l}_{i=1}\phi^S_i-\frac{(j-1)Z}{n'}\right)$ and $A_2=x_{k_{j-1}}\left(\sum^{k_{j-1}}_{i=1}\phi_i-\frac{(j-1)Z}{n'}\right)$. Finally let $\hat{A}_3=x^S_{i_u}\left(\sum^{i_u}_{i=1}\phi^S_i-\frac{jZ}{n'}\right)$ and $A_3=x_{k_{j}}\left(\sum^{k_{j}}_{i=1}\phi_i-\frac{jZ}{n'}\right)$. Then we can bound (\ref{equation:totalerror}) by,
$$
\abs*{\tilde{x}_j^S-\tilde{x}_j} \leq \frac{n'}{Z}\left(\abs*{A_1-\hat{A}_1}+\abs*{A_2-\hat{A}_2}+\abs*{\hat{A}_3-A_3}\right).
$$
Start by bounding,
\begin{equation}
\begin{split}
\abs*{A_1-\hat{A}_1} = \abs*{\left(\sum^{k_j}_{i=k_{j-1}}\phi_ix_i\right)-\left(\sum^{i_u}_{i=i_{l}}\xi_i\right)} &= \abs*{\sum^{k_j}_{i=1}\phi_ix_i -\sum^{k_{j-1}-1}_{i=1}\phi_ix_i-\sum^{i_u}_{i=1}\xi_i+\sum^{i_l-1}_{i=1}\xi_i}\\
\quad &\leq \abs*{\sum^{k_j}_{i=1}\phi_ix_i-\sum^{i_u}_{i=1}\xi_i}+\abs*{\sum^{i_l-1}_{i=1}\xi_i-\sum^{k_{j-1}-1}_{i=1}\phi_ix_i}\\
\quad &\leq \abs*{\sum^{\hat{k}_j}_{i=k_j}\phi_ix_i} + \abs*{\sum^{\hat{k}_{j-1}}_{i=k_{j-1}}\phi_ix_i}\\
\quad &\leq \epsilon Z x^S_{i_u} + \epsilon Z x^S_{i_l} = \epsilon Z(x^S_{i_u}+x^S_{i_l}).
\end{split}
\end{equation}
Here we use the fact that $x^S_{i_u}\geq x_{k_j}$ and that $x^S_{i_{l}}\geq x_{k_{j-1}}$.  Now we bound,
\begin{equation}
\abs*{A_2-\hat{A}_2} = \abs*{x_{k_{j-1}}\left(\sum^{k_{j-1}}_{i=1}\phi_i-\frac{(j-1)Z}{n'}\right) - x^S_{i_l}\left(\sum^{i_l}_{i=1}\phi^S_i-\frac{(j-1)Z}{n'}\right)}.
\end{equation}
Note that the first term is always $\geq 0$, but less than or equal to the second term, since $x^S_{i_l} \geq x_{k_{j-1}}$ and $$
\left(\sum^{i_l}_{i=1}\phi^S_i-\frac{(j-1)Z}{n'}\right) \geq \left(\sum^{k_{j-1}}_{i=1}\phi_i-\frac{(j-1)Z}{n'}\right).
$$
Therefore if $\phi^S_{i_l} > \epsilon Z$ we have that $\abs*{A_2-\hat{A}_2}=0$, and if $\phi^S_{i_l} \leq \epsilon Z$,
\begin{equation}
\abs*{A_2-\hat{A}_2} \leq \abs*{x^S_{i_l}\left(\sum^{i_l}_{i=1}\phi^S_i-\frac{(j-1)Z}{n'}\right)} \leq x^S_{i_l} \phi^S_{i_l} \leq x^S_{i_l}\epsilon Z.
\end{equation}
Finally, we bound,
\begin{equation}
\abs*{\hat{A}_3-A_3} = \abs*{x^S_{i_u}\left(\sum^{i_u}_{i=1}\phi_i^S-\frac{jZ}{n'}\right)-x_{k_{j}}\left(\sum^{k_{j}}_{i=1}\phi_i-\frac{jZ}{n'}\right)}.
\end{equation}
Similarly to the bound $\abs*{A_2-\hat{A}_2}$, this can be bounded by $ \abs*{\hat{A}_3-A_3} \leq x^S_{i_u}\epsilon Z$. Overall we have,
$$
\abs*{\tilde{x}_j-\tilde{x}_j^S} \leq \abs*{A-\hat{A}} \leq \frac{2n'\epsilon Z}{Z}(x^S_{i_l}+x^S_{i_u}) = 2 n' \epsilon(x^S_{i_l}+x^S_{i_u}).
$$
If we set $\epsilon=\alpha/n'$, for $\alpha \geq 0$, then
$$\abs*{\tilde{x}_j-\tilde{x}_j^S} \leq 4 \alpha x^S_{i_u}.
$$
If $\sum^{i_u}_{l=1}\phi^S_l < (j+1)Z/n'$, then by noting that $k_{j+1}=\text{arg}\min_l \big\{\sum^l_{i=1}\phi_i \geq (j+1)Z/n'\big\}$, we have that $x^S_{i_u} < x_{k_{j+1}}$. If $\sum^{i_u}_{l=1}\phi^S_l \geq (j+1)Z/n'$ then $\phi^S_{i_u} \geq Z/n' \geq \epsilon Z$. In this case, we have that $x^S_{i_u} = x_{k_{j+1}}$. Therefore,
$$\abs*{\tilde{x}_j-\tilde{x}_j^S} \leq 4 \alpha x_{k_{j+1}}.
$$

\qed

\end{appendix}

\end{document}

%% file: intro.tex
Sensor networks are being increasingly used in civil engineering
applications. The data from these sensor networks are being used to
monitor and detect changes in the behaviour of instrumented physical structures \citep{Cawley}. 
In this study we are concerned with monitoring and detecting changes in data from a sensor network
installed on a pedestrian bridge. The sensor network consists of two
different types of sensor: accelerometers and strain gauges, that
measure the vertical deflection of the bridge. Both types of
sensors record measurements at a high frequency. These measurements
are recorded without end, which will inevitably pose computational
storage and runtime issues \citep{Bao}. Data compression has therefore been increasingly utilised in literature involving instrumented infrastructures \citep{Khoa, Bao1, Bose}. Despite the computational challenges associated with the sensor data acquisition, it is of interest to study the observed measurements during pedestrian-events, such as a person walking over the bridge, so that we can monitor how the response of the structure to this particular excitement changes over time.

We seek to develop a streaming method that is able to summarise the data corresponding to these pedestrian-events, whilst representing the data obtained from the sensors in a compressed
form. The compressed version of the data will therefore retain
features of the original data, that are prescribed by the user to be \textit{relevant}. This is challenging for a
number of reasons. First, constructing a sequential algorithm (for data compression) in the streaming data regime that can
update at data-acquisition rate is difficult \citep{Lau}; for possibly indefinite data streams this means that such analyzes need to be incrementally updated rather than re-computed every time new data is observed. Second, determining which features
in the original data are important so that they are retained in the
compressed version requires expert knowledge.
%Embodying this expert
%knowledge such that importance can be quantified is not
%straightforward.
Embodying this expert knowledge so that relevant points of the data, with respect to user prescription, are preserved in the compressed version is not straightforward.

In this paper we propose a novel streaming method that summarises data in a compressed
form, whilst preserving data relevant and corresponding to pedestrian-events. The developed method is based on \textit{segmenting} the
time-series -- breaking the time-series up into varying-length
parts. The segmentation (time) points in our method are determined by a
\textit{relevance} score \citep{Moniz, Torgo}. This relevance score quantifies the
importance of each data point in the time-series. We use two types of
relevance score: one that is based only on the data and another
that uses a \textit{query shape}. Using a query shape allows for
particular features in the data to be preserved in the compressed version of the data. The segmentation points can then be used to compress the
time series e.g.\ by using a piecewise linear function between the
segmentation points.

Segmentation is a commonly used method to compress time-series
\citep{Keogh,Fu}. Typically, the segmentation points are computed using
dynamic programming which minimises an error between the original
series and the compressed version \citep{Terzi}. In our method, we develop a
segmentation method for streaming applications -- where segmentation
is done as the time-series is observed in real-time. The proposed
method uses optimal transport \citep{Villani} and linear programming to compute the segmentation points. Moreover, the notion of finding patterns and features in
time-series, as is done here using the relevance score, has received
much attention \citep{Cassisi, Keogh, Keogh2}. Our proposed method
forms a bridge between this notion and data compression by finding a segmentation
of a time-series that is probabilistically optimal with respect to
representing these features. This allows us to compress the sensor data from the aforementioned pedestrian bridge, whilst retaining relevant data corresponding to pedestrian-events.

The remainder of this paper is organised as follows. Section
\ref{sec:data} provides details of the sensor network installed on the pedestrian footbridge that we study, the
accelerometers and strain sensors and the data obtained from both of them. In Section
\ref{sec:relevancefunctions}, we introduce relevance
scores that are used to weight each data point according to their
importance. Sections \ref{sec:time-seri-segm} and \ref{sec:streamingdata} introduce the
segmentation methodology using the relevance score designed for
the sensor data. Section \ref{sec:numer-demonstr} reports a
simulation study designed to gauge the performance of the
method. Further, the methodology is deployed against the sensor
network data.

%% file: motivation.tex
\begin{figure}[!htb]
  \centering
  \minipage{\textwidth}
  \centering
    \includegraphics[width=0.7\linewidth]{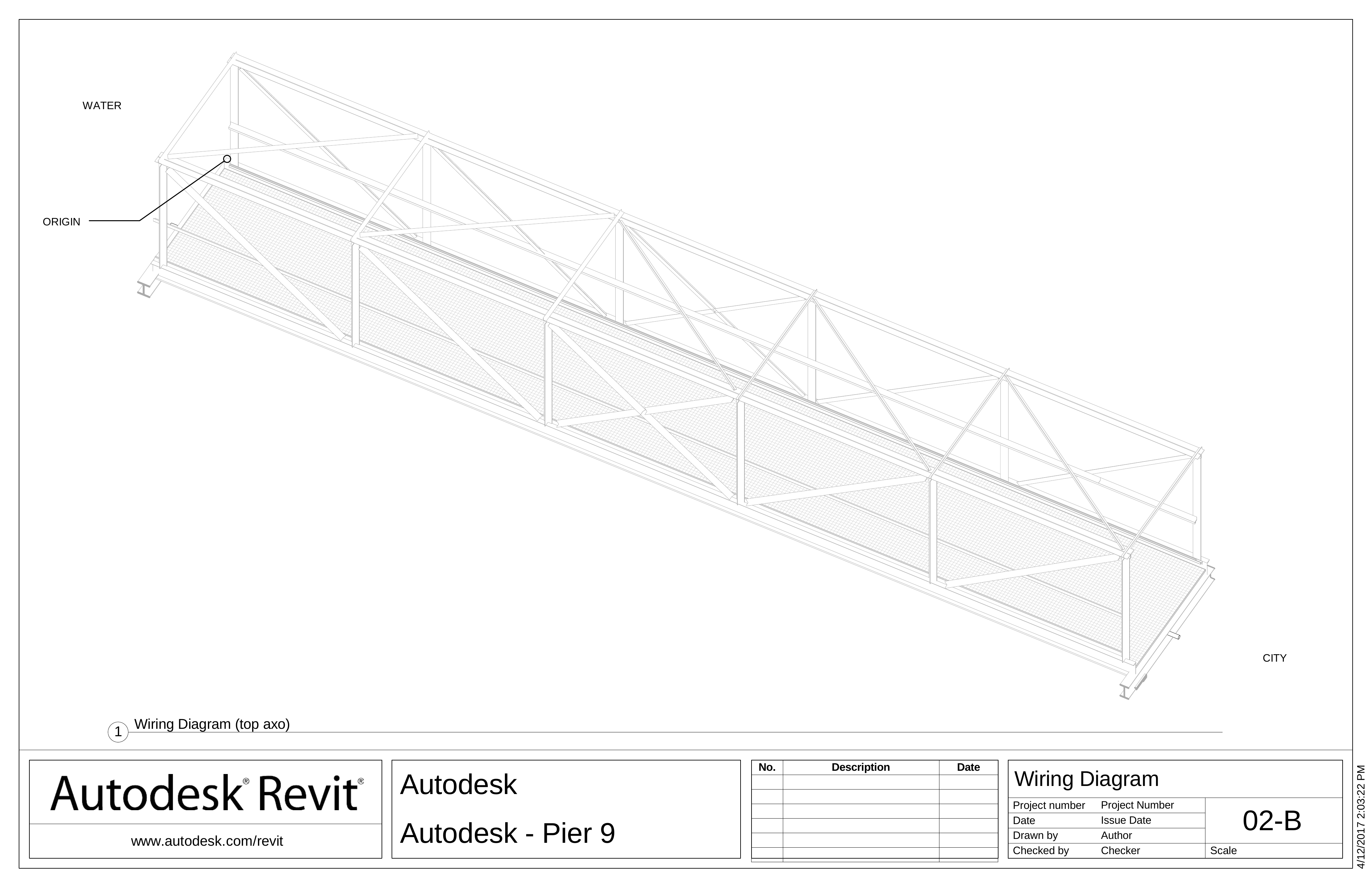}
    \caption{Pier 9 bridge schematic.}
    \label{fig:bridge_schematic}
  \endminipage
\end{figure}
\begin{figure}[!htb]
  \centering
  \minipage{0.48\textwidth}
  \centering
    \includegraphics[width=0.7\linewidth]{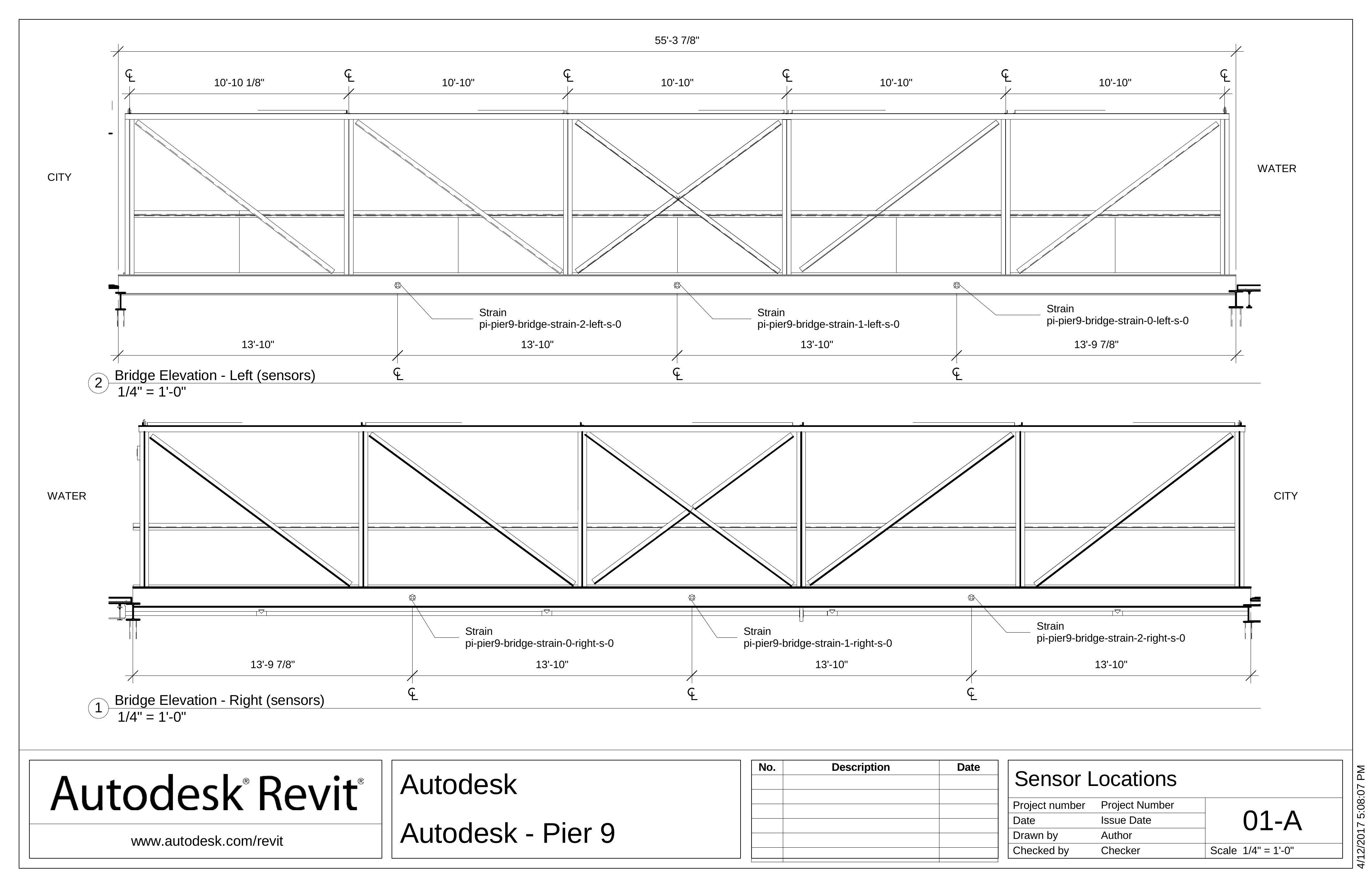}
    \caption{Position of strain gauges.}
    \label{fig:bridge_strain_layout}
  \endminipage\hfill
  \minipage{0.48\textwidth}
  \centering
    \includegraphics[width=0.7\linewidth]{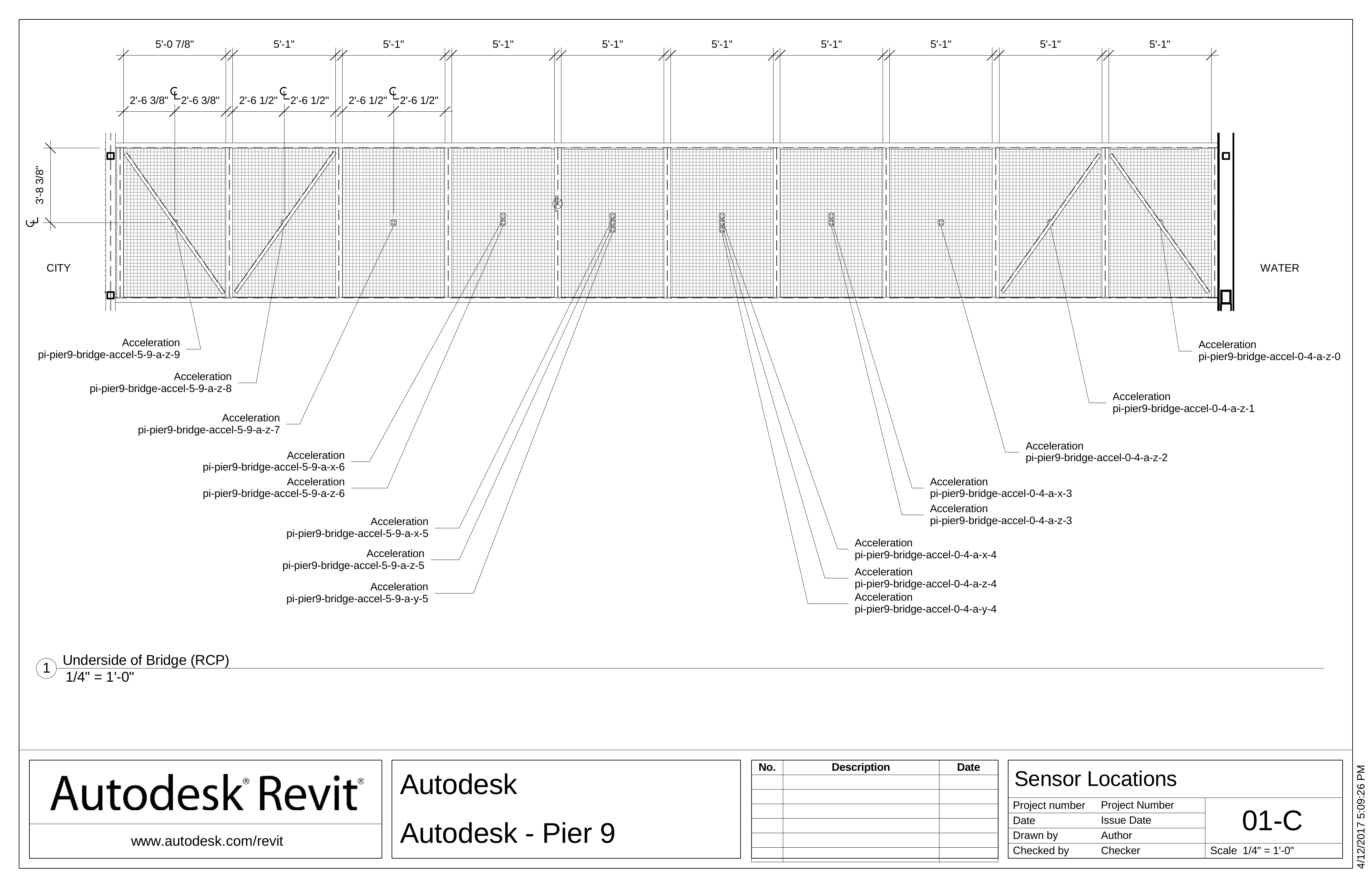}
    \caption{Layout of accelerometers.}
    \label{fig:bridge_accel_layout}
  \endminipage
  \end{figure}

The monitoring of civil infrastructure has typically been performed over finite periods of time. Engineering issues such as fatigue, corrosion and general degradation of concrete and other materials can require long term studies. Due to storage limitations, health monitoring periods tend to be significantly less than the lifetime of the structure. However, certain projects require more continuous monitoring to help ensure safety and to study new materials in the environment. To prepare for this form of continuous monitoring and to provide a testbed for the development of new algorithms, we outfitted an existing indoor pedestrian footbridge with a variety of sensors including accelerometers and strain gauges. The bridge serves as a walkway over a machine shop in a building in San Francisco (Figure \ref{fig:bridge_schematic}) and has a span of 55' and a width of 88.75". The design is a common steel truss bridge. To monitor strain, we installed foil strain gauges at the midpoint of the primary structural element of the bridge, as well as at half the distance between the middle and the ends as seen in Figure \ref{fig:bridge_strain_layout}. The strain gauges were set up to monitor the bending of the primary structural elements of the bridge in a standard Wheatstone bridge configuration using 2 extra gauges for temperature compensation. To keep costs low, custom hardware was designed for the 24 bit Analog to Digital converter (ADC) as a cape for a raspberry pi single board computer which provided the primary interface to the ADC. The core ADC chip used was the Texas Instruments ADS1231 and is capable of supporting 80 samples per second with a $\pm 19.5$mV range. Accelerometers were manufactured by Analog Devices with a sensitivity of $\pm 3$g and wired to a separate 10 bit ADC cape for raspberry pi. Accelerometers were placed in the geometric center of each deck plate section, spaced approximately 5' 1" apart as seen in Figure \ref{fig:bridge_accel_layout}. No filtering or smoothing was performed for any of the sensors resulting in only raw voltage readings stored to a remote cloud based time-series database. Accelerometers were sampled at 40 Hz while strain gauges were sampled at 80 Hz. All devices were synchronized using SNTP (Simple Network Time Protocol) and time-stamps were generated by the Raspbian operating system through a Python script at the time of sampling.

% TODO - move around, shrink/arrange

%This section will introduce the motivating example of infrastructure instrumented with sensors monitoring the condition and performance of it. Two types of sensors will be introduced: strain sensors and accelerometers. Both sensors are instrumented on a pedestrian footbridge .... {\color{red}Specifications!} 

\begin{figure}[!htb]
\centering
\minipage{0.48\textwidth}
\centering
  \includegraphics[width=\linewidth]{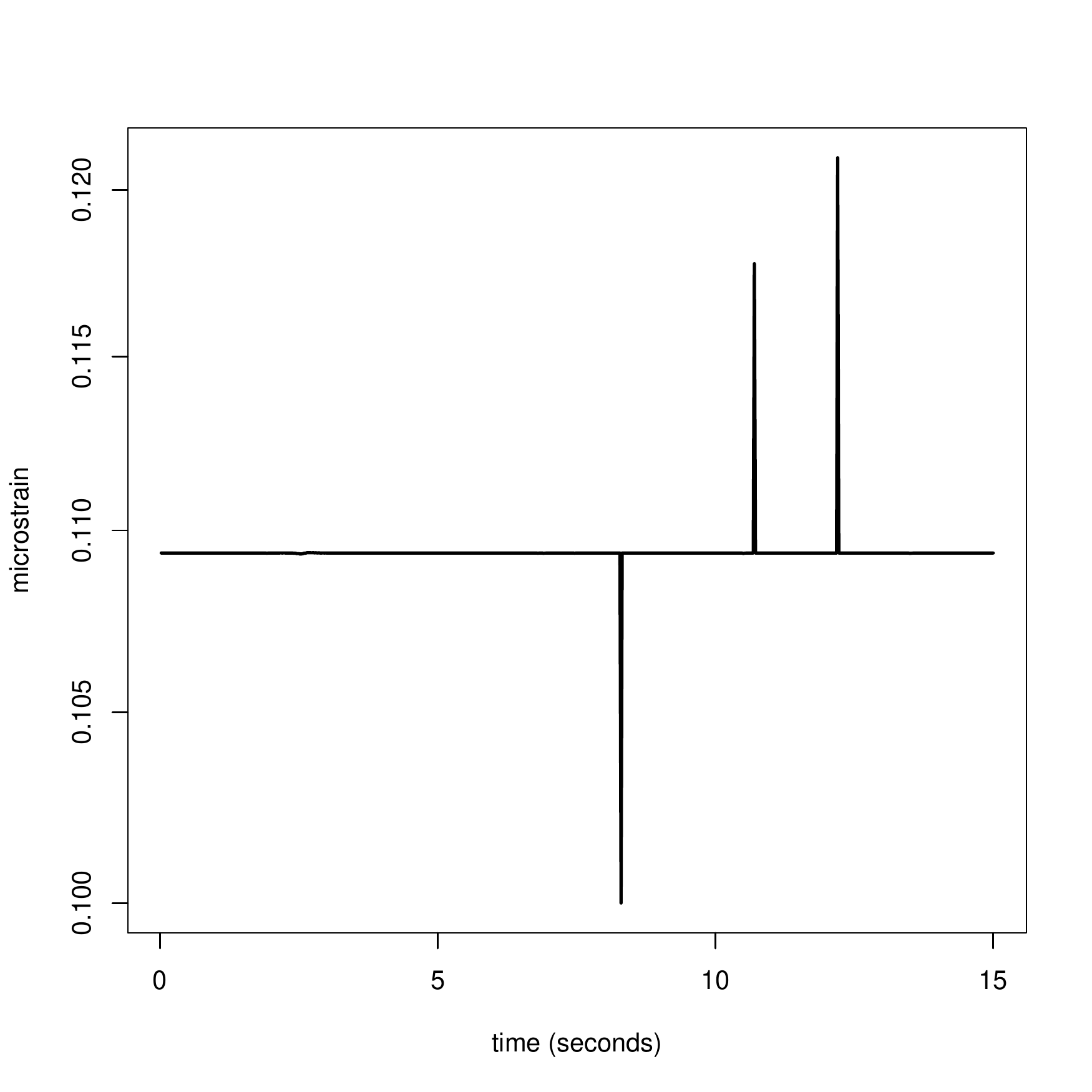}
  \caption{Measurements for a strain sensor installed on the described pedestrian footbridge over time.}\label{fig:ot_signal_1}
\endminipage\hfill
\minipage{0.48\textwidth}
\centering
  \includegraphics[width=\linewidth]{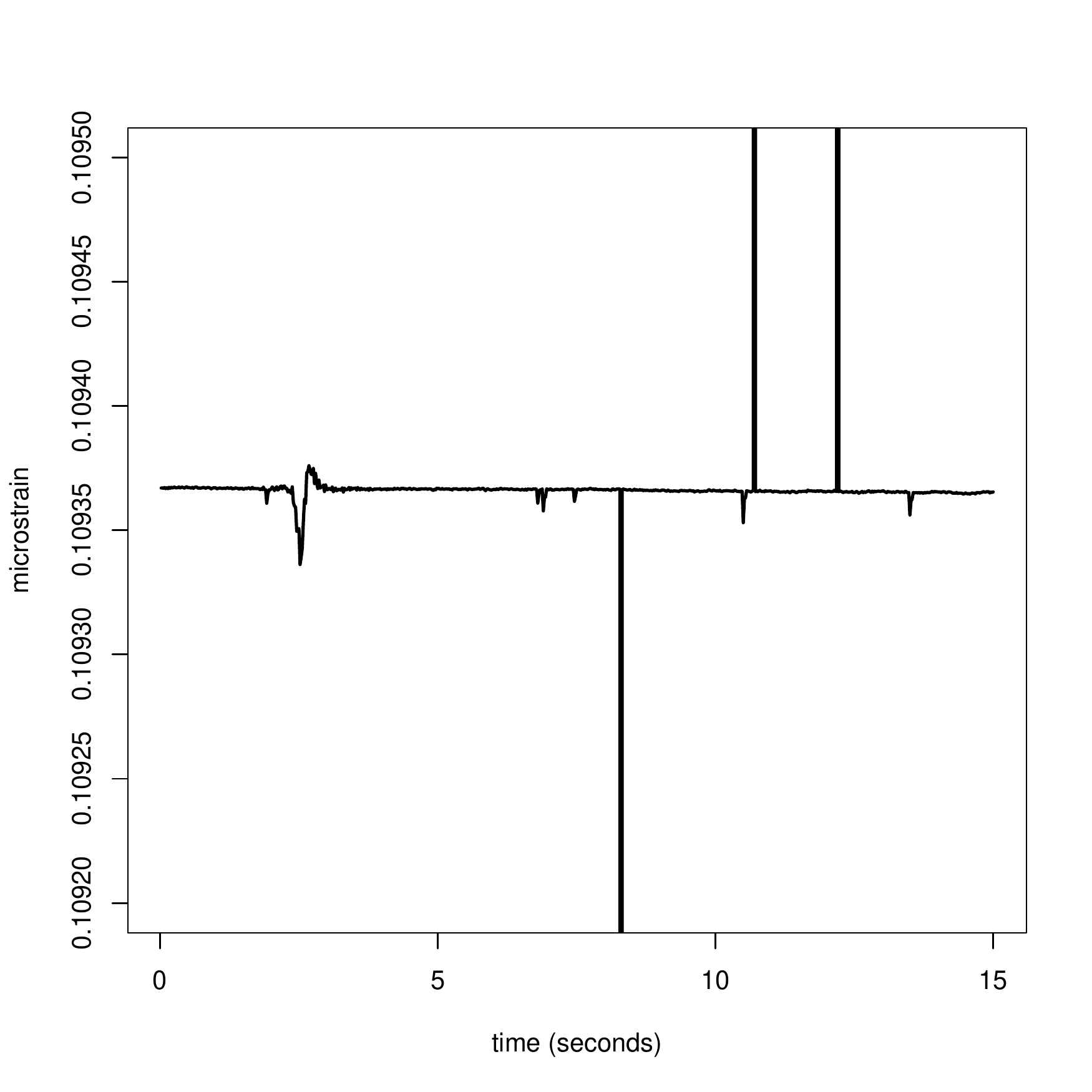}
  \caption{The strain measurements as shown in Figure \ref{fig:ot_signal_1} only with a truncated \textit{y}-scale.}\label{fig:ot_signal_2}
\endminipage
\end{figure}

\begin{figure}[!htb]
\centering
\minipage{0.48\textwidth}
\centering
  \includegraphics[width=\linewidth]{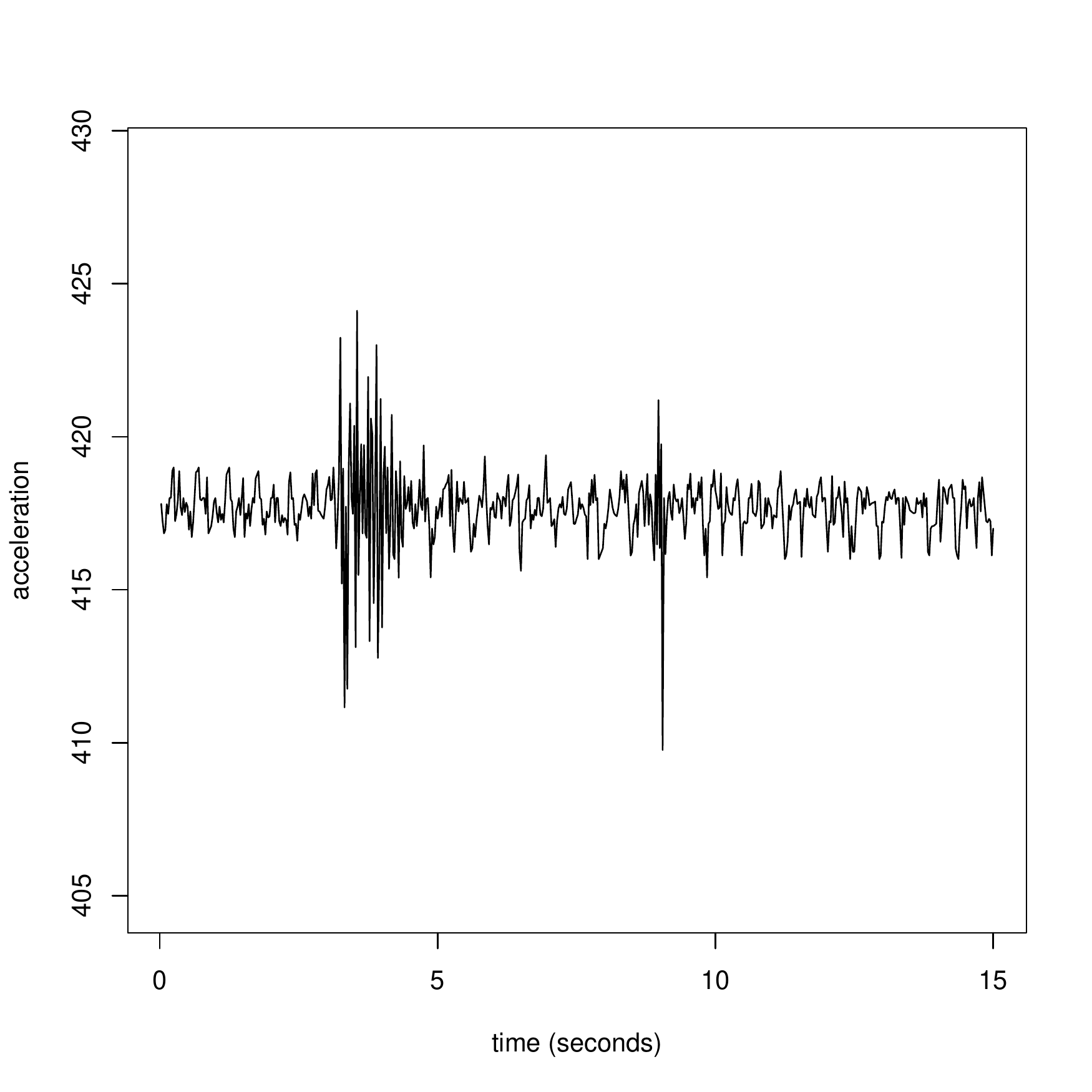}
  \caption{Measurements for an accelerometer installed on the described pedestrian footbridge over time.}\label{fig:ot_acc_signal_1}
\endminipage\hfill
\minipage{0.48\textwidth}
\centering
  \includegraphics[width=\linewidth]{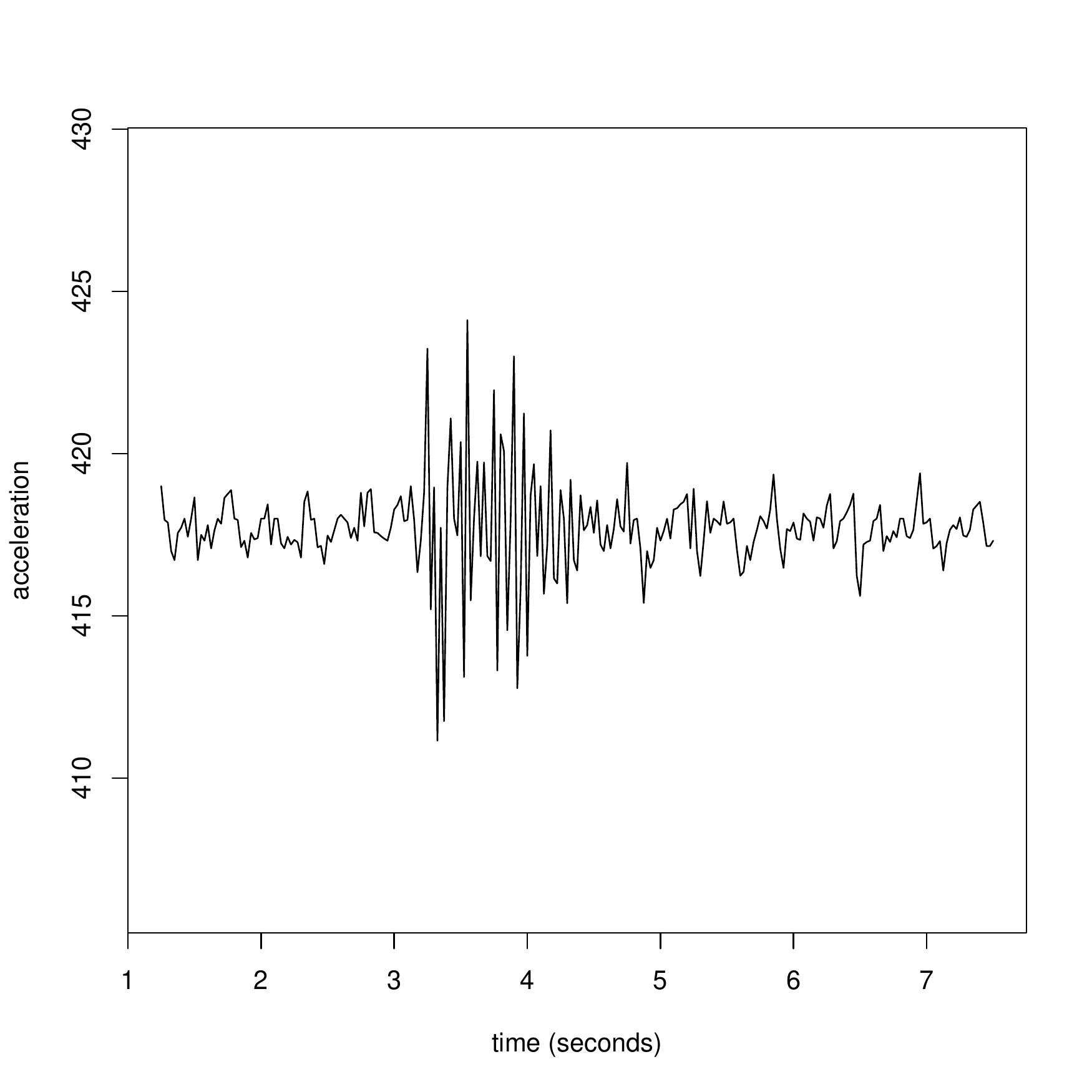}
  \caption{A segment of the acceleration measurements shown in Figure \ref{fig:ot_acc_signal_1}, including a signal that represents a pedestrian-event.}\label{fig:ot_acc_signal_2}
\endminipage
\end{figure}

Figure \ref{fig:ot_signal_1} shows a snippet of time-series data from a single strain sensor (City-side / left: pi-pier9-bridge-strain-2-left-s-0). The same measurements are shown in Figure \ref{fig:ot_signal_2} only with a truncated \textit{y}-scale. Due to environmental electrical noise in the machine shop, initial autoscaled plots are dominated by numerous short outliers with high amplitude. However for the application of using this data to monitor pedestrian-events (such as a person walking) on the bridge, what an analyzer of such strain sensor data is concerned with are the small sinusoidal-like signals towards the start of the illustrated time-series. This represents the structural response of the bridge as a pedestrian-event occurs. It is the periods of data that contain recurrent loading, the traversal of the pedestrian's interaction with the bridge, that are of interest for our application.
%In the case of a standard compression technique that aims to minimise reconstruction loss (e.g. dynamic programming), this would likely be left out of the reconstruction due to the large number and high magnitude of the outliers. The technique proposed in this paper is concerned with maintaining such a wave in a lower dimensional form by compressing everything else.
Next, we consider the snippet of time-series data shown in Figure \ref{fig:ot_acc_signal_1}; this time-series shows measurements from an accelerometer (City-side: pi-pier9-bridge-accel-5-9-a-z-9). Figure \ref{fig:ot_acc_signal_2} shows a segment of this time-series that contains a visibly high magnitude oscillatory-like signal, that represents a pedestrian-event occurring near the accelerometer. This particular pattern of data obtained from an accelerometer is therefore of interest to an analyzer when aiming to monitor the response of the bridge to pedestrian-events over time. Any compressed version of these two data-sets should aim to preserve relevant signals such as the ones discussed here, in order to effectively be able to monitor the pedestrian-events to a similar degree as one can do with the raw data shown here. The next section explains how one can weight which data points within the time-series obtained from the accelerometers and strain sensors considered in this section are relevant, with respect to being able to monitor these pedestrian-events.

%Just as we would like a reconstruction from a compression to preserve the sinusoidal wave representing a signal from the strain sensor described above, we would like preserve these such signals in the time-series obtained by the accelerometer in any constructed compression.

%It's also distribution-free invariant to shifts and scaling in the time-series, as we are picking out key features, and may not be using simply raw values!

%The remainder of the paper is structured as follows. The next section explains the segmentation problem for time-series compression. In Sec. \ref{sec:relevancefunctions}, the concept of relevance functions for sweighting points in time-series with respect to key features or patterns is considered. Following this, a segmentation method based on this relevance function and optimal transport is proposed in Sec. \ref{sec:otsegmentation}, and the benefits of it are detailed, including the implementation speed. The next section outlines an approximation to the segmentation obtained in Sec. \ref{sec:otsegmentation} for the streaming data regime. Finally in Sec. \ref{sec:numericalexperiments}, some numerical simulations with strain sensor and accelerometer data are shown.

%%% Local Variables: 
%%% mode: latex
%%% TeX-master: "optimal_transport_compression"
%%% End:

%% file: relevance.tex
In this section we introduce the notion of relevance scores. A
relevance score operates on the time-series and is used to
characterise the importance of each data point. Two types of relevance
scores are selected for use with the accelerometer and strain
data. Consider the univariate real-valued time series
$y_1,y_2,\dots,y_n$ observed at the time-stamps $x_1<x_2<\dots< x_n$
respectively. A relevance score operates on the time-series,
transforming each value into a real-value \textit{score} that
quantifies its importance; the higher the score, the more important the data point. Denote the relevance score for a
time-series $y_1,y_2,\dots,y_n$ as $\phi_1,\phi_2,\dots,\phi_n$.
We now introduce two different types of relevance scores.

% This section explores using relevance functions to characterise the importance of points in a time-series; important points can then be preserved in a compressed form using the accompanying segmentation algorithm proposed in Sec \ref{sec:otsegmentation}. The idea of these relevance functions $\phi$ was introduced in \cite{Moniz} and \cite{Torgo}. Consider the time-series $\big\{(x_i,y_i)\big\}_{i=1}^{n}$ (where $x_i$'s are the time-stamps), and we shall assume henceforth that $x_i < x_{i+1}$. Define a \textbf{relevance function} $\phi_i:=\phi_i(\textbf{y})$ to be the function which transforms the values $\textbf{y}:=(y_1,\ldots,y_n)$ that form the time-series $\big\{(x_i,y_i)\big\}_{i=1}^{n}$, into scores of importance $\boldsymbol \phi:=(\phi_1,\ldots,\phi_n)$. There are two main classes of relevance functions that we will consider the remainder of this paper.

\bigskip
\noindent \textbf{Data-driven relevance scores}\\
\smallskip
Define the following relevance scores, based on only the data in the time-series $y_1,y_2,\ldots,y_n$,
\begin{itemize}
\item[(i)] $\phi_i=\lvert y_i\rvert^p,\quad p\in\mathbb{Z}\geq1$,
\item[(ii)] $\phi_i=\lvert y_{i}-y_{i-1}\rvert^p, \quad y_{0}=0$,
\item[(iii)] $\phi_i=H(\lvert y_i - y_{i-1}\rvert -\beta)^p ,\quad \beta >0.$,
\end{itemize}
where $H$ is the Heaviside function.
% These relevance functions above were also utilised in \cite{Liu}.
% The form of the relevance scores shown above depend on only one or
% two values within the vector $\boldsymbol \phi$ for the computation of
% $\phi_i$.
%Choosing a relevance function $\phi$ that reflects these features is the crux of the proposed methodology.
The various relevance scores capture different features of the
time-series. In (i), large scores are associated with high magnitude
values. In (ii), large magnitude differences in the contiguous pairs
of the time series lead to high relevance scores. In (iii), the scores
at instance $i$ satisfying $\lvert y_{i}-y_{i-1}\rvert >\beta$ will
have a non-zero value otherwise the score is zero. This represents a procedure
where only large differences between contiguous pairs of values lead to
a non-zero score.  The higher the value of $p$ in these relevance scores, the greater the difference between the score for relevant and non-relevant points. The scores (i) and (ii) were used in \cite{Liu}.
% In (i), the
% greater the relevance $\phi_i$ is for larger absolute values of
% centred\xx{why centred?} values of $y_i$. For (ii), the higher the
% value $p$ takes, the greater the relevance $\phi_i$ is for larger
% absolute values of $y_{i}-y_{i-1}$. Finally for (iii), only values of
% $i$ satisfying $\lvert y_{i}-y_{i-1}\rvert >\beta$ will have a
% non-zero value of $\phi_i$. 
% One of these types of relevance function, (ii), is used later in the
% paper alongside data obtained from the accelerometers introduced in
% the previous section. This function is used to highlight
% large-oscillatory signals in the accelerometer data, periods of
% interest and relevance to an analyzer.
For the accelerometer data we shall use the relevance scores presented
in (ii) with $p=2$. This choice is motivated by the
oscillatory features in the data as noted in Section \ref{sec:data}.

\bigskip

\noindent \textbf{Query-based relevance scores} \\
\smallskip 
Another type of relevance score can take a \textit{query} shape as an
additional input, which captures a feature of the original data that
we wish to retain in the compressed version of the data. A query shape is described by the real-valued vector
$\boldsymbol q = (q_1,\dots,q_m)$, with $m$ odd. One such score is
% This type of relevance function $\phi$ captures pattern- and shape-based features in the data that one would like to preserve in the reconstruction. In this case, $\phi_i$ could be a distance between a subsequence (with center $x_i$) and a prescribed query. For more information on this turn to the work in \cite{Esling} for a background on the field of pattern recognition in time-series data. On this note, consider the relevance function,
\begin{equation}
\phi_i = \vert\vert \boldsymbol q - \boldsymbol y_{(i-\gamma):(i+\gamma)}
\vert\vert^{-2p},
\label{equation:distancetransformfunc}
\end{equation}
where $\gamma=(m-1)/2$, also $\boldsymbol y_{(i-\gamma):(i+\gamma)}=(y_{i-\gamma},\ldots,y_{i+\gamma})$ and $\vert\vert\cdot\vert\vert$ is the Euclidean norm. 
% where $\big\{q_i\big\}_{i=1}^{m}$ is a prescribed query shape that we
% would like to find the time-series.
The relevance score in \eqref{equation:distancetransformfunc} is
used for the strain sensor data so that the sinusoidal-wave type shape seen in the previous section
is preserved in the compressed data. For scale and shift invariance, there are warping
distances \citep{Keogh3, Paparrizos} that can be used instead of the
Euclidean metric used above; this is sufficient for the scope of this
paper. When $x_i$ denotes the center of a subsequence of the
time-series that matches the query shape well, the greater the
importance associated with the value $y_i$ through the relevance
score. 

\smallskip 

\begin{figure}[!htb]
\centering
\minipage{0.45\textwidth}
\centering
  \includegraphics[width=\linewidth]{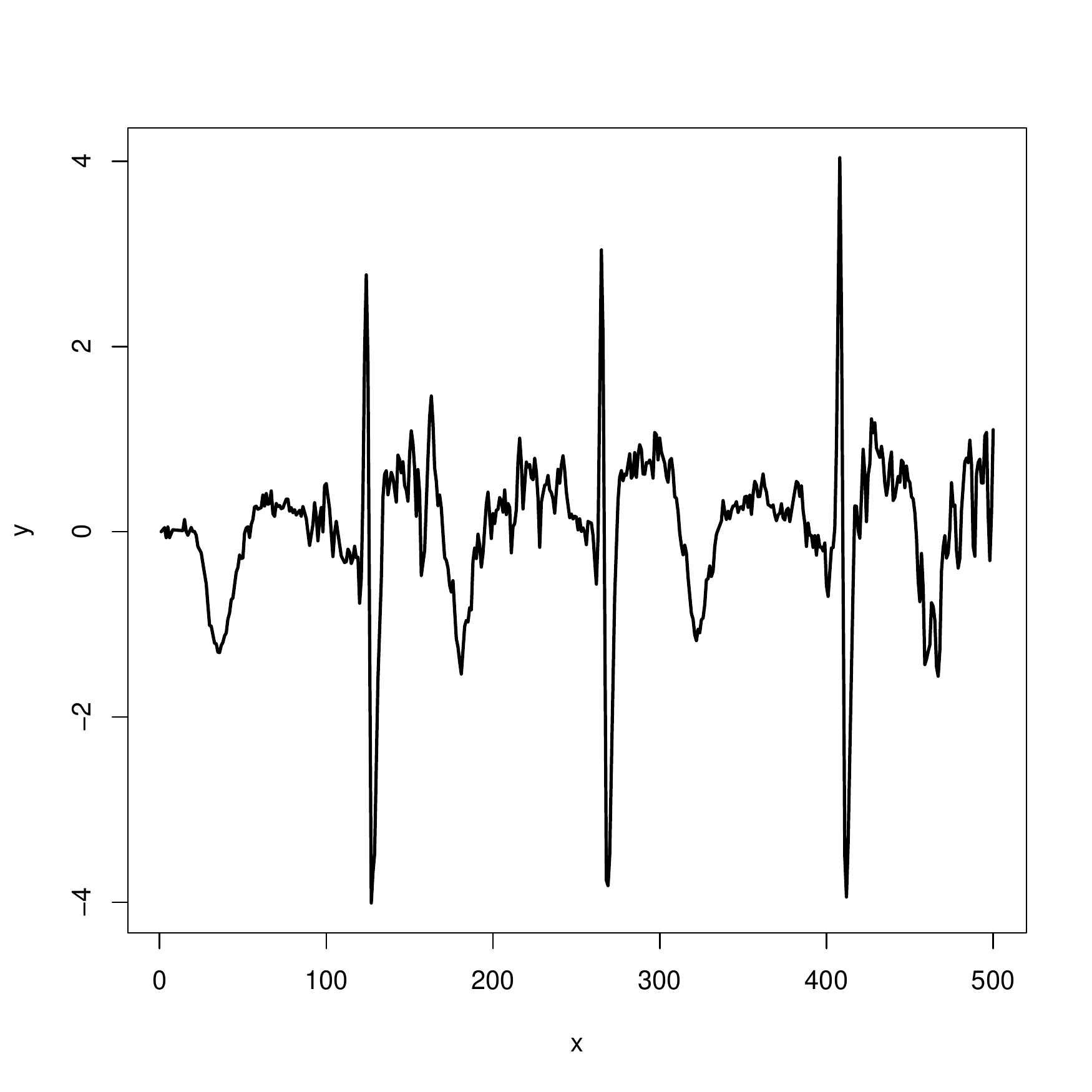}
  \caption{An ECG signal.}\label{fig:ecg}
\endminipage\hfill
\minipage{0.45\textwidth}
\centering
  \includegraphics[width=\linewidth]{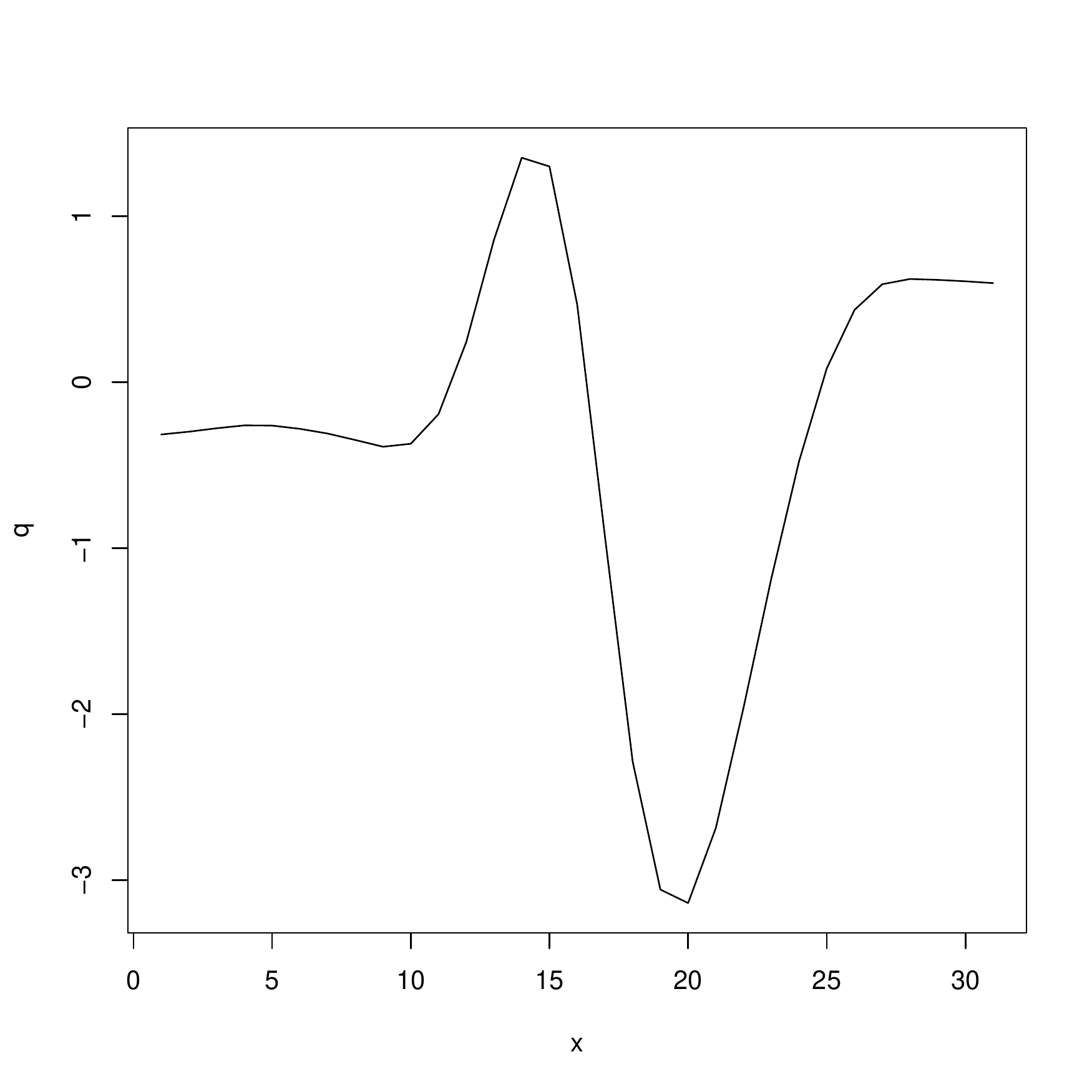}
  \caption{The query shape $q_i$, for $i=1,\ldots,31$.}\label{fig:query}
\endminipage \vfill
\minipage{0.45\textwidth}
\centering
  \includegraphics[width=\linewidth]{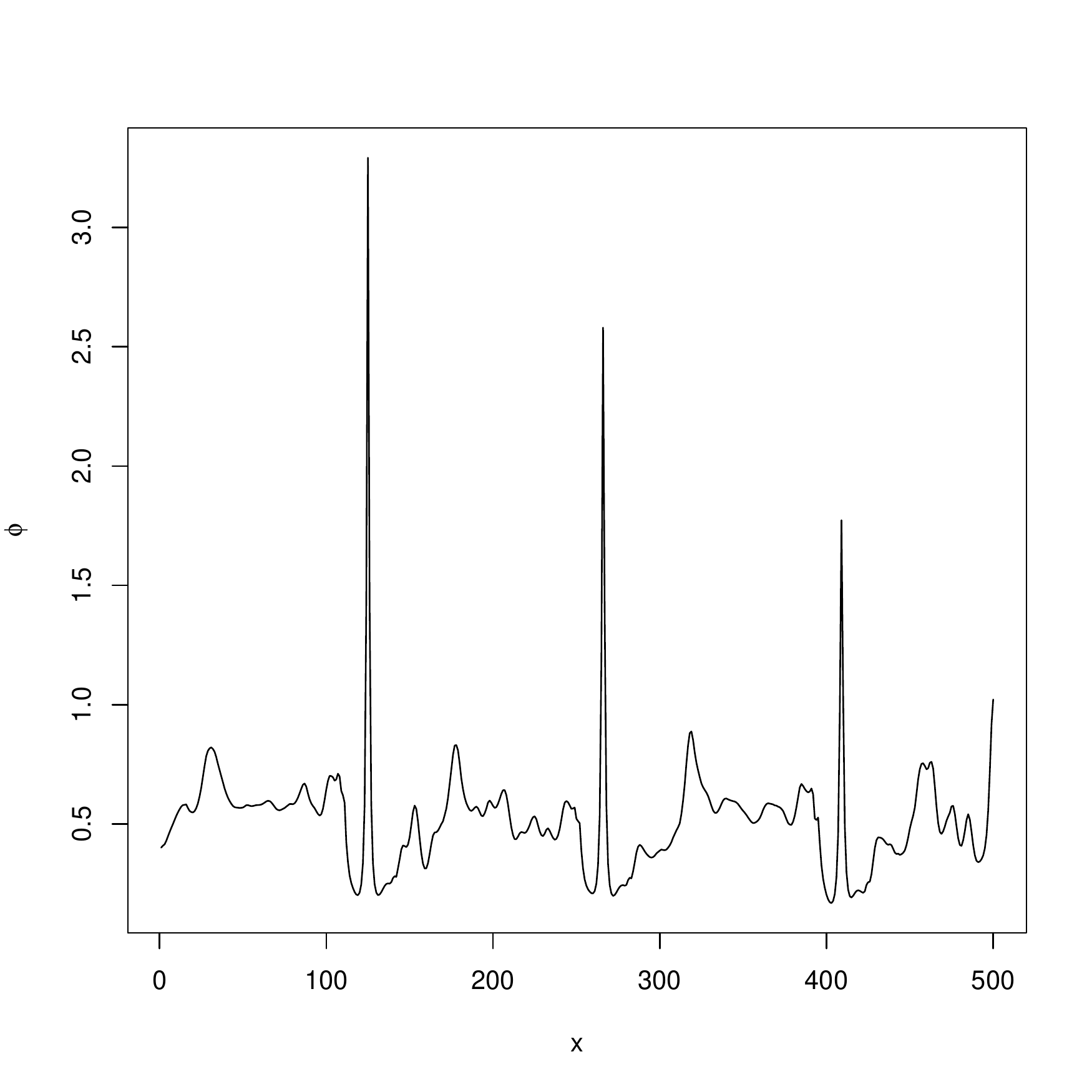}
  \caption{The relevance score $\phi_i$ in (\ref{equation:distancetransformfunc}), for $i=1,\ldots,500$, describing the similarity between the query shape and each subsequence of the original time-series.}\label{fig:transform_func}
  \endminipage\hfill
\minipage{0.45\textwidth}
\centering
  \includegraphics[width=\linewidth]{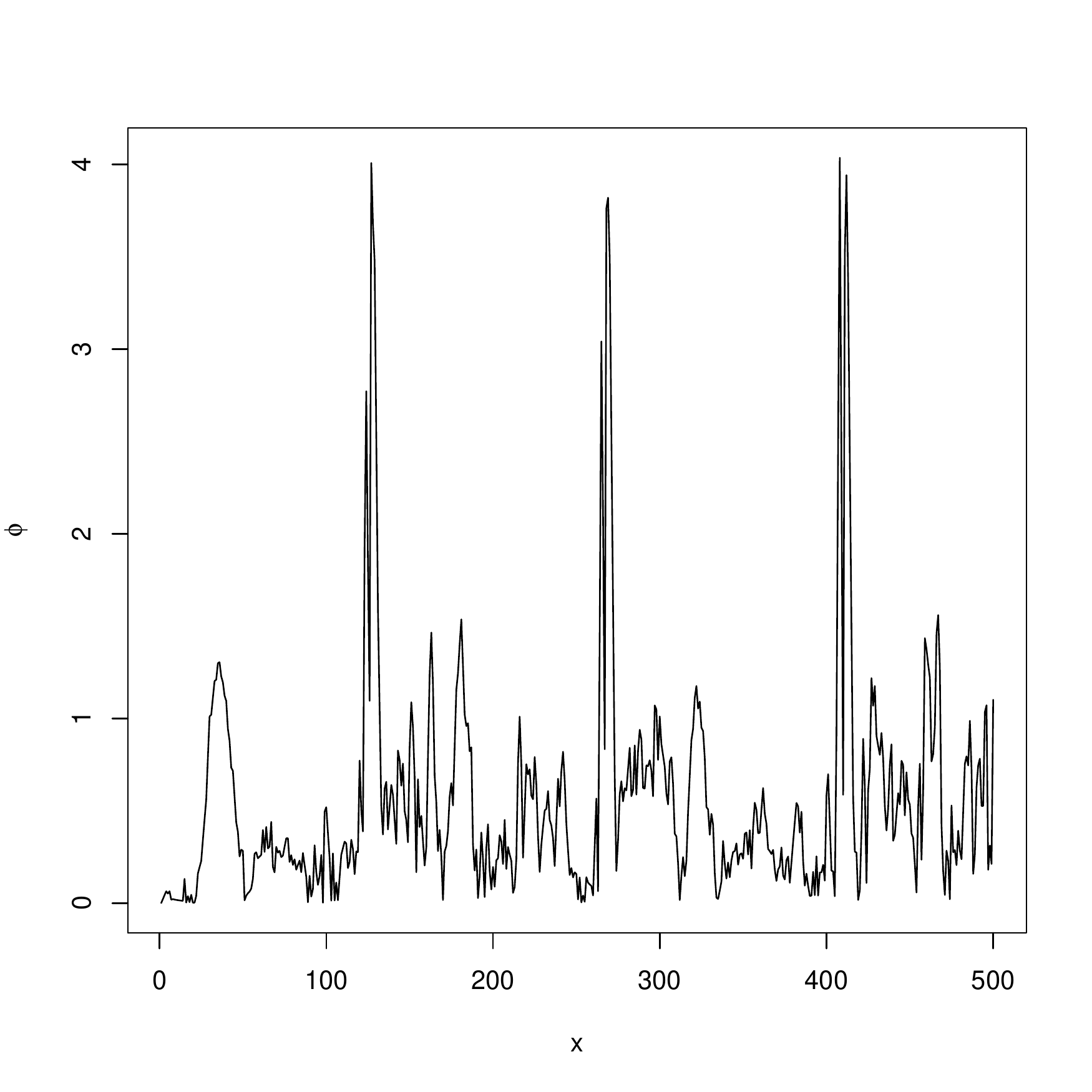}
  \caption{The relevance score $\phi_i$ in (ii), for $i=1,\ldots,500$, describing the absolute magnitude of each point in the original time-series.}\label{fig:transform_func_absolute}
  \endminipage
\end{figure}

The following example will illustrate the behaviour of both types of
relevance score. Figure \ref{fig:ecg} shows an example of an ECG
time-series of length $500$. The query shape, of length $31$, is
displayed in Figure \ref{fig:query}. Figure
\ref{fig:transform_func} shows the relevance scores in
(\ref{equation:distancetransformfunc}) for the time-series, with $p=1$. Note that the
relevance score is higher for subsequences of the original ECG
time-series that match the query shape well, e.g. during the three
occurences of peak signal. On the other hand, Figure \ref{fig:transform_func_absolute} shows the relevance scores in (ii) for the time-series, with $p=1$. Due to the high magnitude of the peak signals relative to the rest of the time-series, these relevance scores are similar to those in Figure \ref{fig:transform_func}, but exhibit slightly more variance.
The next section will introduce a method that divides a time-series
based on its relevance scores. This \textit{segmentation} of the
time-series is used to generate a compressed version of the data.

%% file: segmentation.tex
\iffalse

% FOR INTRODUCTION

When analyzing sensor data from instrumented infrastructure for the application of structural health monitoring, it is important to analyze this data in a rapid manner as typically the data is acquired at a fast rate. Such analyzes might include outlier detection, similarity search or fitting parametric models to the data. A landmark study in \cite{Keogh2} had great success in implementing dimensionality reduction for the application of similarity search on time-series. Compression-based sampling of time-series data obtained from accelerometers attached to beam structures was found to have a positive effect on the management of data for the application of structural health monitoring in \cite{Bao}.

\fi

\label{sec:otsegmentation}

Relevance scores, introduced in Section \ref{sec:relevancefunctions},
characterise the desired features in a time series that are of
interest. This section introduces a method that divides the
time-series based on its relevance scores into \textit{segmentation}
(time) points. Interpolation between these segmentation time points
leads to a compressed version of the time-series. First, Section \ref{sec:segm-time-seri-comp} describes how the segmentation points
are computed in the static case where the time-series is not a data stream. When streaming data is considered, the method to divide the time-series into segmentation points is required to be recursive, as it assumed infeasible to re-compute this segmentation after every point is added to the time-series. Section
\ref{sec:streamingdata} therefore introduces a method to incrementally compute an approximation to the segmentation points that one would obtain by following the static methodology in Section \ref{sec:segm-time-seri-comp}, for use in the streaming data setting.

% Now that relevance functions have been introduced, describing the importance of points in a time-series $\big\{(x_i,y_i)\big\}_{i=1}^{n}$, the compression methodology will be proposed in this section. First, we outline segmentation for time-series data, and how it can be used for compression. Then we will explain a method to segment time-series data using the relevance functions introduced in the previous section. Following that, in Sec. \ref{sec:streamingdata} we consider how this method can be implemented in the streaming data regime; this is important when considering data obtained from contemporary sensing technologies.

\subsection{Segmentation of time-series}\label{sec:segm-time-seri}

The segmentation of a time-series leads to time-series data compression. As aforementioned, this compression is important for data acquired by sensor networks fitted to instrumented infrastructure, in order to reduce the complexity and increase the efficiency of any analysis. Segmentation can be used for time-series data compression by breaking the original time-series up into segments; one can then reconstruct a compressed version of the original time-series using these segments alongside some interpolation method. The type of compressed reconstructions that this paper considers are known as piecewise aggregate approximations (PAA) \citep{Fu}.
%These segments could take the form of different time-series states or symbols, e.g. in symbolic aggregate approximation (SAX), or piecewise functions, e.g. in piecewise aggregate approximation (PAA). This study will be concerned with the latter of these two, however it should be noted that the methodology outlined here is aimed at finding suitable segment breakpoints in the time-series rather than proposing particular representations of the time-series segments.
For the time-series $y_1,y_2,\ldots,y_n$ observed at the time-stamps $x_1,x_2,\ldots,x_n$, we denote $n'$ \textit{segmentation points} by $\tilde{x}_1,\tilde{x}_2,\ldots,\tilde{x}_{n'}$, where $n' \leq n$ and $\tilde{x}_j \in \big\{x_i\big\}_{i=1}^{n}$, for all $j=1,\ldots,n'$. Since $n' \leq n$, the data is compressed. These points define a unique approximation $S\big(x_l;\big\{\tilde{x}_j\big\}_{j=1}^{n'}\big)$, which is a compressed reconstructed version of the original time-series point $y_l$, for $x_l \in \big\{x_{i}\big\}_{i=1}^{n}$. Exact forms for a couple of these compressed reconstructions are given in Sec. \ref{sec:analysisreconstruction}. A common metric to assess the space-efficiency of the compression of the original time-series is the \textit{compression ratio}, given by,
$$
C_{R}=\frac{n}{n'}, \quad \text{with } C_{R} \geq 1 .
$$
In practice, a segmentation algorithm will typically be implemented and evaluated by \textbf{(a)} specifying a desired compression ratio, and then reporting the error of the approximation, or \textbf{(b)} specifying a condition for the segmentation to satisfy (e.g. maximum approximation error), and then reporting the compression ratio. The methodology in this paper is inline with \textbf{(a)}. A naive choice for the segmentation points would simply be $n'$ evenly spaced points; however for relatively small values of $n'$ this would lead to important signals in the data being lost in the compression. Therefore the segmentation problem is concerned with choosing the points $\tilde{x}_1,\tilde{x}_2,\ldots,\tilde{x}_{n'}$ for some predefined objective, such as minimizing the approximation error. For example, our predefined objective here is to preserve key, relevant features in the compression. Dynamic programming can be implemented to find the particular segmentation which minimizes the total error of the reconstruction away from the original data; this is at the computational expense of $\mathcal{O}(n^2 n')$ \citep{Terzi}.
%Whilst dynamic programming concentrates on the reconstruction from the compression and not on the specific points in the time-series data,
In the next two sections we propose a segmentation algorithm that will focus on points in the original time-series data that have high relevance to an analyzer.

\subsection{Computing the segmentation points} \label{sec:segm-time-seri-comp}

The contribution of this work, a methodology for time-series segmentation whilst preserving relevant features of the original data, is now described in this section for the case where the time-series $y_1,y_2,\ldots,y_n$ is not being streamed. We consider the opposite case in the Sec. \ref{sec:streamingdata}. We seek a segmentation of a time-series that is constructed using the segmentation points $\tilde{x}_1,\tilde{x}_2, \ldots, \tilde{x}_{n'}$, where $1 \leq n' \leq n$. Recall that the original time-series points were assigned relevance scores in Sec. \ref{sec:relevancefunctions} that described their importance. These can be made into weights via the normalization, $w_i=\phi_i/(\sum^{n}_{j=1}\phi_j)$. The segmentation points $\tilde{x}_1,\tilde{x}_2,\ldots,\tilde{x}_{n'}$ can also be assigned weights, and we would like each of them to be as equally relevant in the compression. Therefore, let $\tilde{w}_{1},\tilde{w}_2,\ldots,\tilde{w}_{n'}$, where $\tilde{w}_j=1/n'$ for $j=1,\ldots,n'$, be these weights.

We will now describe the method used to compute the segmentation points $\tilde{x}_1,\tilde{x}_2,\ldots,\tilde{x}_{n'}$, based on the method of optimal transport \citep{Villani}. At first glance, it seems reasonable to resample the $n'$ segmentation points, using any standard resampling method \citep{Douc}, from the weighted time-series points. This approach is not ideal for the objectives in this paper, since there is little guarantees using these resampling methods of a particular placement for a segmentation point. Instead in this paper we use a deterministic linear transformation from the original time-stamps to the segmentation points. It will be shown later in this section and in Sec. \ref{sec:analysisreconstruction} that this transformation allows one to guarantee particular placements of the segmentation points and prove properties about the corresponding compressed reconstruction. Define the \textit{coupling matrix} $\eta \in \mathbb{R}^{n \times n'}$, with the constraints,
\begin{equation}
\sum^{n}_{i=1}\eta_{i,j}=\tilde{w}_j, \quad \sum^{n'}_{j=1}\eta_{i,j}=w_i.
\label{equation:constraints}
\end{equation}
Using the coupling matrix, segmentation points $\tilde{x}_1,\tilde{x}_2,\ldots,\tilde{x}_{n'}$ can be computed for this methodology via the following linear transformation,
\begin{equation}
\tilde{x}_j = \sum_{i=1}^{n}\eta_{i,j}n'x_{i},
\label{equation:segmentationpointtransform}
\end{equation}
for $j=1,\ldots,n'$. We are interested in the particular coupling matrix, known as the optimal coupling, that solves the well-known Monge-Kantorovitch optimization problem,
\begin{equation}
\text{arg}\min_{\eta \in  \mathbb{R}^{n \times n'}} \left(\sum^n_{i=1}\sum^{n'}_{j=1}\eta_{i,j}(x_i-\tilde{x}_j)^2\right).
\end{equation}
%that minimises the expected distance over the space of all couplings $\mathcal{E}$, $\mathbb{E}_{\eta}[(X-\tilde{X})^2]$, with respect to the coupling $\eta$ is the solution to the well-known Monge-Kantorovitch problem: the $\mathbb{L}_{2}$ Wasserstein distance,
%$$
%W_{\mathbb{L}_2}\left\{(\textbf{x},\textbf{w}),(\tilde{\textbf{x}},\tilde{\textbf{w}})\right\}=\text{arg} \min_{\eta \in \mathcal{E}} \sqrt{\mathbb{E}_{\eta}[(X-\tilde{X})^2]}.
%$$
%Minimising the distance $\mathbb{E}_{\eta}[(X-\tilde{X})^2]$ is equivalent to maximising the covariance, with respect to $\eta$, between $X$ and $\tilde{X}$ via
%$$
%\mathbb{E}_{\eta}[(X-\tilde{X})^2] = \mathbb{E}[X^2]+\mathbb{E}[\tilde{X}^2]-2\mathbb{E}[X]\mathbb{E}[\tilde{X}]-2\mathbb{C}_{\eta}[X,\tilde{X}].
%$$
In our case, the scheme chooses segmentation points that are as close to $x_1, x_2,\ldots,x_{n}$ as possible whilst satisfying the constraints in (\ref{equation:constraints}).
%It is through this fact, that one can see what this type of segmentation achieves: DISCUSS HOW IT ALLOWS US TO ACHIEVE FAST, FLEXIBLE COMPRESSIONS BECAUSE WE MADE MASS-BASED CONSTRAINTS.
%Note that
%$$
%\sum^{\tilde{x}_{j+1}}_{i=\tilde{x}_j+1}w_i=\frac{1}{n'}.
%$$
Linear programming can be used to numerically compute the optimal coupling $\eta$, at the computational expense of $\mathcal{O}(n\log n)$. The matrix $\eta$ will have at most $n+n'+1$ non-zero elements. The pseudocode of this algorithm is given in Algorithm \ref{alg:general} in Appendix \ref{sec:algolp}. Note that the time-stamps $x_1,x_2,\ldots,x_n$ are not an input for this algorithm; this is because all time-stamps are assumed ordered. The segmentation scheme does not use information about the reconstruction, and only which features of the original time-series the reconstruction would be required to preserve. This is why it can utilise linear programming to solve the problem, and hence become significantly cheaper than alternative segmentation methods. If it is not acceptable in a particular application for the segmentation points to not necessarily take integer values, then we can use $\tilde{x}_j=\ceil*{\sum^n_{i=1}\eta_{i,j}n'x_i}$, for $j=1,\ldots,n'$, instead of (\ref{equation:segmentationpointtransform}).
%This segmentation is deterministic, however it means that the segmentation points are no longer neccessarily equal to one of the time-stamps $x_1,x_2,\ldots,x_n$. If this is not acceptable in a particular application, then one can compute the segmentation point $\tilde{x}_j$, for $j=1,\ldots,n'$, by sampling one of $x_{k}$, for $k=1,\ldots,n$, with probability $\eta_{k,j}n'$. Note that by using this random method, there is a non-zero probability of a segmentation point $\tilde{x}_j$ being equal to any time-stamp $x_1,x_2,\ldots,x_n$ as long as $w_j>0$.
%In this case we have the consistency property that there will be segmentation points on every element in $\textbf{x}$ as $n' \to \infty$, i.e. $\textbf{x} \subset \tilde{\textbf{x}}$ as $n' \to \infty$.
The procedure in this section was proposed as a resampling scheme for non-parameteric data-assimilation in \cite{Reich}.
%For the application presented in this paper, this random method is not used; instead henceforth we will assume that the linear transformation in (\ref{equation:segmentationpointtransform}) is always used to compute the segmentation points.
The constraints in (\ref{equation:constraints}), that dictate the form that the optimal coupling matrix $\eta$ takes, are influenced by the relevance score of the time-series points. By computing the segmentation points $\tilde{x}_1,\tilde{x}_2,\ldots,\tilde{x}_{n'}$ in this section using the coupling matrix $\eta$, we therefore designate more segmentation points towards periods of time-stamps that correspond to time-series points with high relevance scores.

Note that following Algorithm \ref{alg:general} in Appendix \ref{sec:algolp}, each segmentation point falls inside a particular interval, i.e.
\begin{equation} \tilde{x}_j \in [x_{k_{j-1}},x_{k_{j}}], \quad k_u=\text{arg}\min_l \Bigg(\sum^l_{i=1}w_i \geq u/n'\Bigg), \quad u=j-1,j\label{equation:intervals}\end{equation}
for $j=1,\ldots,n'$ and with $k_{0}=1$. This interval forms an important part of the extension of this methodology to the streaming data case considered in Sec. \ref{sec:streamingdata}. The expression in (\ref{equation:intervals}) is also an important aspect of the proposed methodology, as it makes sure that there will be a segmentation point within a certain period of data, even if all the points within it are not particularly relevant. This is useful in many applications where sensor data have long-term drifts in background noise; this guaranteed interval will allow sparsely placed segmentation points to keep track of this drift. The next section will now consider how to reconstruct a compressed version of the original time-series using the segmentation points computed in this section. This reconstruction will therefore preserve highly relevant periods of time-series data in the compression to a greater extent than irrelevant periods of data.

\subsection{Compressed reconstruction}

\label{sec:analysisreconstruction}

This section will explore the PAA \citep{Fu} compressed reconstruction $S\big(t;\big\{\tilde{x}_j\big\}_{j=1}^{n'}\big)$, for $t \in \big\{x_i\big\}_{i=1}^{n}$, of the original time-series $y_1,y_2,\ldots,y_n$ using the segmentation points computed within the methodology presented in the previous section. Let the segmentation points $\tilde{x}_1,\tilde{x}_2,\ldots,\tilde{x}_{n'}$ have the indices $s_1,s_2,\ldots,s_{n'}$ in the original time-series so that $\tilde{x}_j=x_{s_j}$, for $j=1,\ldots,n'$. Two examples of PAA reconstructions are now given: define the piecewise constant approximation,
\begin{equation}
S\big(t;\big\{\tilde{x}_j\big\}_{j=1}^{n'}\big)=\frac{1}{(\tilde{x}_{l+1}-\tilde{x}_{l}+1)}\sum^{s_{l+1}}_{i=s_l}y_i, \quad \text{where } \tilde{x}_{l} \leq t \leq \tilde{x}_{l+1},
\label{equation:piecewiseconstant}
\end{equation}
and piecewise linear approximation,
\begin{equation}
S\big(t;\big\{\tilde{x}_j\big\}_{j=1}^{n'}\big)=y_{s_{l}}+\left(t-\tilde{x}_{l}\right)\left(\frac{y_{s_{l+1}}-y_{s_{l}}}{\tilde{x}_{l+1}-\tilde{x}_{l}}\right), \quad \text{where } \tilde{x}_{l} \leq t \leq \tilde{x}_{l+1},
\label{equation:piecewiselinear}
\end{equation}
for $l=0,\ldots,n'$ and where $\tilde{x}_{0}=1$ and $\tilde{x}_{n'+1}=n$. Another simple alternative to this approximation is the piecewise regression approximation.

The error of the relevance scores of the compressed reconstruction, utilising the segmentation points computed in the previous section, away from the relevance scores of the original time-series is now considered. This error metric is of particular interest to the scope of this paper since the proposed methodology is designed for when the practioner would like to preserve relevant features in the compressed version. We shall assume a piecewise reconstruction satisfying $$S\left(t;\big\{\tilde{x}_j\big\}_{j=1}^{n'}\right) \in \left[\min_{x_i \in [\tilde{x}_j,\tilde{x}_{j+1}]}y_i,\max_{x_i \in [\tilde{x}_j,\tilde{x}_{j+1}]}y_i\right],$$ where $\tilde{x}_j \leq t \leq \tilde{x}_{j+1}$. Recall that $\phi_1,\phi_2,\ldots,\phi_n$ are the relevance scores of the original time-series, and now let $\tilde{\phi}_i$ be the relevance score of $S(x_i;\big\{\tilde{x}_j\big\}_{j=1}^{n'})$ for $i=1,\ldots,n$. Then,
\begin{equation}
\abs*{\tilde{\phi_i}-\phi_i} < \frac{\sum^{n}_{t=1}\phi_t}{n'},
\label{equation:reconstructionerrorbound}
\end{equation}
for all $i = 1,\ldots,n$. The derivation of this is shown in Appendix \ref{sec:showingreconstruction}.

%%% Local Variables: 
%%% mode: latex
%%% TeX-master: "optimal_transport_compression"
%%% End:

%% file: streaming_segmentation.tex
This section will now remove the assumption made in the previous section that the time-series is not streamed. In the streaming data case, new data points are added to the time-series $y_1,y_2,\ldots,y_n$ sequentially, possibly indefinitely.
%Considering this type of data is important with the increase in modern sensing and monitoring systems for instrumented infrastructure; sensors can acquire data continuously on the structure's current performance or health for real-time condition monitoring \citep{Lau}.
%Given that this data is streamed potentially for an indefinite amount of time, it is of interest to the analyzer to update analyzes of the data (such as computing the compression proposed in the previous section) on-the-fly rather than re-computing every time a new element is added to the time-series.
%In addition to this, it is preferable for all of the data not to be stored for analyzes due to space-memory constraints of the user and therefore only a synopsis of the data is kept. This is a less important aspect of a streaming segmentation algorithm than the runtime issue, given that using segmentation for compression which has the primary purpose of reducing the dimension of the time-series.
%One could of course simply implement the linear programming algorithm in Algorithm \ref{alg:general} after every element is added to the time-series, however we assume that for indefinite series the runtime associated with this would be infeasible.
%A technique that could also be used to deal with streaming data is to implement the required analyzes on the previous $m$ time-series points, thus creating a sliding window.
We propose a recursive approximation to the segmentation points that one would obtain from using the methodology presented in the previous section; this approximation is updated every time a new data point is added to the time-series. An approximation is required since the segmentation points are derived using the linear transformation in (\ref{equation:segmentationpointtransform}) and this transformation is affected by the constraints in (\ref{equation:constraints}). These constraints are dependent on the normalized weights $w_i$, for $i=1,\ldots,n$; each time a data point is added to the time-series all previous weights will change, leading to the position of all segmentation points changing. On another note, since the approximation is recursive, it is more efficient than re-computing the segmentation points using the methodology in the previous section each time the time-series is added to. There are two aspects to this approximation that are discussed in this section. First, we explain how one can update the number of segmentation points used for the compression $n'$ as the time-series increases in size. Second, we outline how we approximate the $n'$ segmentation points; a user-defined approximation error controls how accurate one wants the approximation to be.
 %that would be obtained from using the methodology in the previous section on the entire time-series $\big\{(x_i,y_i)\big\}_{i=1}^{n}$. Therefore we trade a bounded error on the segmentation points for the ability to be able to quickly compute the compression at any point in time. The user can specify the desired approximation error; this prescribed error can be reduced if one was to know they had a large amount of time to compute the compression or if data was streamed in at an infrequent rate. 
A general outline of the approximation is given below:
\begin{enumerate}
\item[(1)] Initialize the algorithm by observing the first $n$ points of the time-series, setting $L \leftarrow n$, $n'$ to be the user's choice, $\Delta Z \leftarrow  0$ and $Z\leftarrow \sum^{n}_{l=1}\phi_l$. Also initialize:
\begin{itemize}
\item $\boldsymbol x^S = (x_1^S,\ldots,x_L^S)=(x_1,\ldots,x_n)$,
\item $\boldsymbol \phi^S = (\phi^S_1,\ldots,\phi^S_L) = (\phi_1,\ldots,\phi_n)$,
\item$\boldsymbol \xi = (\xi_1,\ldots,\xi_L)=(\phi_1x_1,\ldots,\phi_n x_n)$.
\end{itemize} 
\item[(2)] Prescribe a user-defined level of accuracy $\alpha\in[0,1]$ and set $\epsilon\leftarrow\alpha/n'$. 
\item[(3)] Set $n \leftarrow n + 1$, and observe the new data point $y_{n}$ in time-series (or $\boldsymbol y_{(n-\gamma):(n+\gamma)}$ using a buffer, if required for a query-driven relevance score), at the time-stamp $x_{n}$, and compute associated relevance score $\phi_{n}$. Set $\boldsymbol x^S = (\boldsymbol x^S , x_n)$, $\boldsymbol \phi^S = (\boldsymbol \phi^S, \phi_n)$ and $\boldsymbol \xi = (\boldsymbol \xi, \phi_n x_n)$. 
\item[(4)] \textbf{Update $\boldsymbol n'$ and prune $\boldsymbol x^S$, $\boldsymbol \phi^S$ and $\boldsymbol \xi$:} Set $\Delta Z \leftarrow \Delta Z + \phi_{n}$. If:
\begin{equation}
\Delta Z \geq Z / n',
\label{equation:conditionsegpoints}
\end{equation}
then implement:
\begin{itemize}
\item[(4i)] Set $n' \leftarrow n' + 1$, $\epsilon\leftarrow\alpha/n'$, $Z \leftarrow Z + \Delta Z$ and $\Delta Z \leftarrow 0$.
\item[(4ii)] Set $z \leftarrow L$ and while $z \geq 3$, implement:
\begin{itemize}
\item If $\phi^S_z > \epsilon \sum^{L}_{l=1}\phi_l^S$, then set $z \leftarrow z - 1$. On the other hand, if $\phi^S_z \leq \epsilon \sum^{L}_{l=1}\phi_l^S$, then compute
\begin{equation}
i = \text{arg} \min_{j\in[2,z]}\left(\sum^{z}_{l=j}\phi^S_l \leq \epsilon \sum^{L}_{l=1}\phi_l^S\right),
\label{equation:indexcondition}
\end{equation}
and set $\boldsymbol x^S = (x^S_1,\ldots,x^S_{i-1},x^S_z,\ldots,x^S_L)$, $\boldsymbol \phi^S = (\phi^S_1,\ldots,\phi_{i-1}^S,\sum^{z}_{l=i}\phi^S_l,\phi^S_{z+1},\ldots,\phi^S_L)$ and $\boldsymbol \xi = (\xi_1,\ldots,\xi_{i-1},\sum^{z}_{l=i}\xi_l,\xi_{z+1},\ldots,\xi_L)$. Set $L \leftarrow \abs*{\boldsymbol \phi^S}$, and $z \leftarrow i-1$.
\end{itemize}
\end{itemize}
\item[(5)] For $j=1,\ldots,n'$ implement:
\begin{itemize}
\item[(5i)] \textbf{Update approximation to end-points of interval in (\ref{equation:intervals}):} Set $$i_l = \text{arg} \min_{l}\left(\sum^{l}_{i=1}\phi_i^S \geq (j-1)\sum^{L}_{i=1}\phi_i^S/n'\right) \quad \text{and} \quad i_u = \text{arg} \min_{l}\left(\sum^{l}_{i=1}\phi_i^S \geq j\sum^{L}_{i=1}\phi_i^S/n'\right).$$
Then $x^S_{i_l}$ is an approximation to $x_{k_{j-1}}$ and $x^S_{i_u}$ is an approximation to $x_{k_{j}}$.
\item[(5ii)] \textbf{Update approximation to segmentation points:} Compute the approximation to the $j$'th segmentation point, $\tilde{x}^S_j=\left(\sum^{i_u}_{i=i_l}\xi_i+E\right)/(Z/n')$, where $E$ is an adjustment term explained in more detail in Appendix \ref{sec:appendixstreamingdata}.
\end{itemize}
%\item[(5)] \textbf{Update approximation to end-points of interval in (\ref{equation:intervals}):} At all times, we update an approximation $\hat{k}_j$ to $k_j$, for $j=1,\ldots, n'$, given in (\ref{equation:intervals}), and the weighted sum $\left(\sum^{\hat{k}_j}_{l=\hat{k}_{j-1}}\phi_l x_l\right)$ with $\hat{k}_{0}=1$.
%\item[(6)] \textbf{Maintain approximation error:} We guarantee that for any index $i=1,\ldots,n$, we have access to an approximation $\hat{i}$ where,
%for $j=1,\ldots,n'$.
%\item[(7)] \textbf{Update approximation to segmentation points:} Use the weighted sum $\left(\sum^{\hat{k}_j}_{l=\hat{k}_{j-1}}\phi_l x_l\right)$ in order to obtain an approximation, $\tilde{x}_j^S$, to the segmentation point $\tilde{x}_j$, for $j=1,\ldots,n'$.
\item[(6)] Return to step (3).
\end{enumerate}
\iffalse
 Update the number of segmentation points used $n'$; a condition for this number to be determined is given in Sec. \ref{sec:approximationsegpoints}. This update is an important aspect of computing segmentations for (indefinitely) streamed time-series.
\item[(4)] For each segmentation point $\tilde{x}_j$, for $j=1,\ldots,n'$, update a bounded approximation to the end-points of the interval $[x_{k_{j-1}},x_{k_j}]$ that contains the segmentation point.
\item[(5)] Using an updated weighted midpoint between the approximated end-points of the interval $[x_{k_{j-1}},x_{k_j}]$, compute an approximation to the segmentation point $\tilde{x}^{S}_j$.
\end{enumerate}
%Each of these steps are now discussed in more detail in the following subsections.

%Another benefit, computational cost grows with only $\mathbb{O}(n'}$, and when $m=\mathcal{O}(n)$, this is a lesser rate than computing optimal transport on every iteration.

%One can also see from numerical simulations that the error of this approximation actually decreases as more elements and segmentation points are added.
\fi

The procedure outlined in the steps above is given in more detail in Appendix \ref{sec:appendixstreamingdata}. The intuition behind the approximation is the following. The vectors $\boldsymbol x^S$, $\boldsymbol \phi^S$ and $\boldsymbol \xi$ keep a synopsis of the time-stamps, relevance scores, and products of time-stamps and relevance scores over the time-series. In step (4ii), some elements of these vectors are combined and summed together when the corresponding relevance scores are low; these elements are unlikely to have segmentation points on them. As this synopsis is pruned over time, generating approximations from the elements of it instead of the entire time-series will be efficient. Now, since we know that the segmentation point $\tilde{x}_j$, for $j=1,\ldots,n'$, is within the interval $[x_{k_{j-1}},x_{k_{j}}]$, it is important to always maintain approximations to the end-points of the interval in (\ref{equation:intervals}); this is done in step (5i).
%The condition in (\ref{equation:indexcondition}) guarantees that we can approximate the index $i$ of any time-stamp $x_i$ by the index $\hat{i}$, with a bounded error on the sum of the relevance scores associated with the data points $y_i,\ldots,y_{\hat{i}}$ (assuming w.l.o.g. that $\hat{i}\geq i$):
%\begin{equation}
%\abs*{\sum^{i}_{l=1}\phi_l - \sum^{\hat{i}}_{l=1}\phi_l} \leq \epsilon \left(\sum^{n}_{l=1}\phi_l\right),
%\label{equation:gcondition}
%\end{equation}
%This condition is guaranteed by using a synopsis of the entire data-set with a prescribed resolution (see Appendix \ref{sec:appendixstreamingdata} for details). Using this we have that the approximation to the end-points satisfies,
Using the condition in (\ref{equation:indexcondition}) we have that approximations to the end-points of the interval in (\ref{equation:intervals}) satisfy,
\begin{equation}
\abs*{\sum^{i_l}_{l=1}\phi_l^S - \sum^{k_{j-1}}_{l=1}\phi_l} \leq \epsilon \left(\sum^{n}_{l=1}\phi_l\right), \quad  
\abs*{\sum^{i_u}_{l=1}\phi_l^S - \sum^{k_j}_{l=1}\phi_l} \leq \epsilon \left(\sum^{n}_{l=1}\phi_l\right).
\label{equation:gcondition}
\end{equation}
A consequence of this on the accuracy of the segmentation point approximation, given by a rolling weighted sum $\sum^{i_u}_{l=i_l}\xi_l$ in step (5ii), is that,
\begin{equation}
\abs*{\tilde{x}^S_j-\tilde{x}_j} \leq 4 \alpha x_{k_{j+1}},
\label{equation:errorboundsegpoints}
\end{equation}
for $j=1,\ldots,n'$ and where $x_{k_{n'+1}}=n$. This bound on the segmentation points is proved in Appendix \ref{sec:proof}.

As an example to see how the number of segmentation points $n'$ is updated in step (4) as data points are added to the time-series, consider the following time-series:
$$
y_1=0, y_2=3, y_3=0 \text{ and } y_4=1 \text{ observed at the time-stamps } x_1=1,x_2=2,x_3=3 \text{ and } x_4=4.
$$
We assume that we start with $n'=2$, and let $\phi_i=\abs{y_i-y_{i-1}}$ with $y_{0}=0$. Therefore, $\phi_1=0,\phi_2=3,\phi_3=3$ and $\phi_4=1$. After the fifth data point, $y_5$, enters the time-series at the time-stamp $x_5=5$, we have that $Z=7$ and $\Delta Z = \phi_5$. If $y_5=5$, we have that $\phi_5=4$ and therefore $\Delta Z > Z/n'$. In this case we would increase the number of segmentation points by one. If on the other hand $y_5=0$, we have that $\phi_5=1$ and therefore $\Delta Z < Z/n'$. In this case the number of segmentation points would stay as $n'=2$. The next section gives a numerical demonstration of this approximation to the segmentation points of a time-series, in the streaming data regime.

%% file: numerics.tex
The following section will demonstrate the methodology presented throughout the paper with the application of the method to simulated streaming data and data from the accelerometers and strain gauges instrumented on the pedestrian footbridge introduced in Sec. \ref{sec:data}. These demonstrations prove the effectiveness of the proposed compression technique for the application of efficiently managing data from instrumented infrastructure,  whilst preserving key features in the original sensor data.

\label{sec:numericalexperiments}

\subsection{Simulated streaming data}

This section will investigate the effectiveness of the streaming data approximation, presented in Sec. \ref{sec:streamingdata}, for the segmentation points obtained from the optimal transport algorithm introduced in Sec. \ref{sec:otsegmentation}. Recall that this approximation is required in the streaming data regime since it is assumed infeasible to re-compute the segmentation points every time a new element is added to the time-series. The implementation of the approximation is described in Appendix \ref{sec:appendixstreamingdata} and segmentation points are added on-the-fly when the condition in (\ref{equation:conditionsegpoints}) is met. The simulated time-series considered in this problem is,
\begin{equation}
y_i=\sin^3(i\pi 200 /(5 \times 10^5)), \quad \text{for } i=1,\ldots,5 \times 10^5
\label{equation:sinusoidal}
\end{equation}
where $x_i=i$, and the relevance score used is $\phi_i=\abs*{y_i}$, with $p=2$. This time-series is chosen to simulate frequent occurrences of a particular magnitude-based feature in the data, that will hopefully allow the segmentation points to shift periodically to the peaks of the sinusoidal waves when they enter into the time-series. Initially we start with $n'=500$, and $n=1000$. The value $\alpha=0.2$ is used. After all elements have been added to the time-series, there are $n'=984$ segmentation points.

\begin{figure}[!htb]
\centering
\minipage{0.48\textwidth}
\centering
  \includegraphics[width=\linewidth]{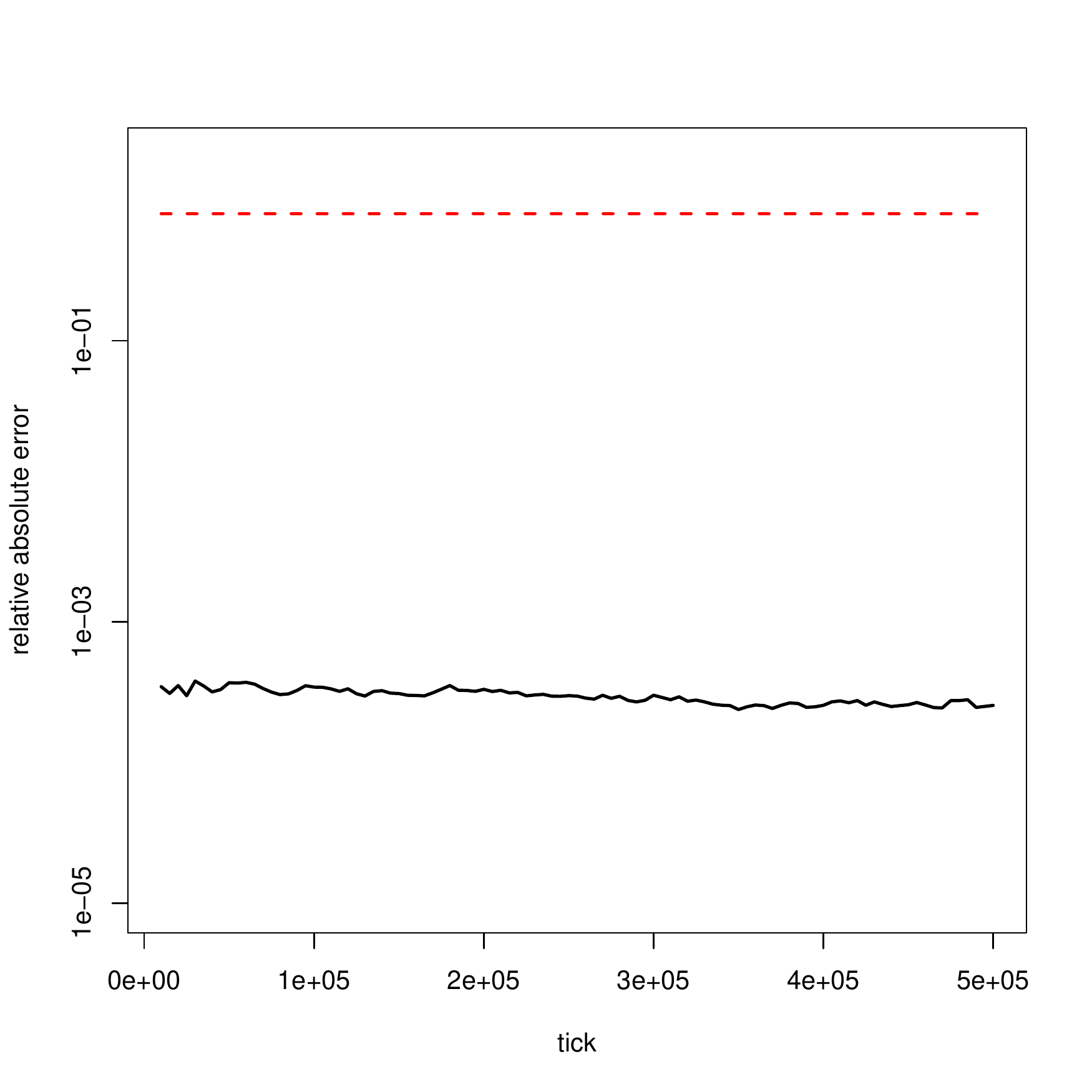}
  \caption{Mean absolute relative error of all approximate segmentation points after every 5000'th element has been added to the time series in (\ref{equation:sinusoidal}). The theoretical bound in (\ref{equation:errorboundsegpoints}) is also shown in the dashed line.}\label{fig:error_over_time_seg_points}
\endminipage\hfill
\minipage{0.48\textwidth}
\centering
  \includegraphics[width=\linewidth]{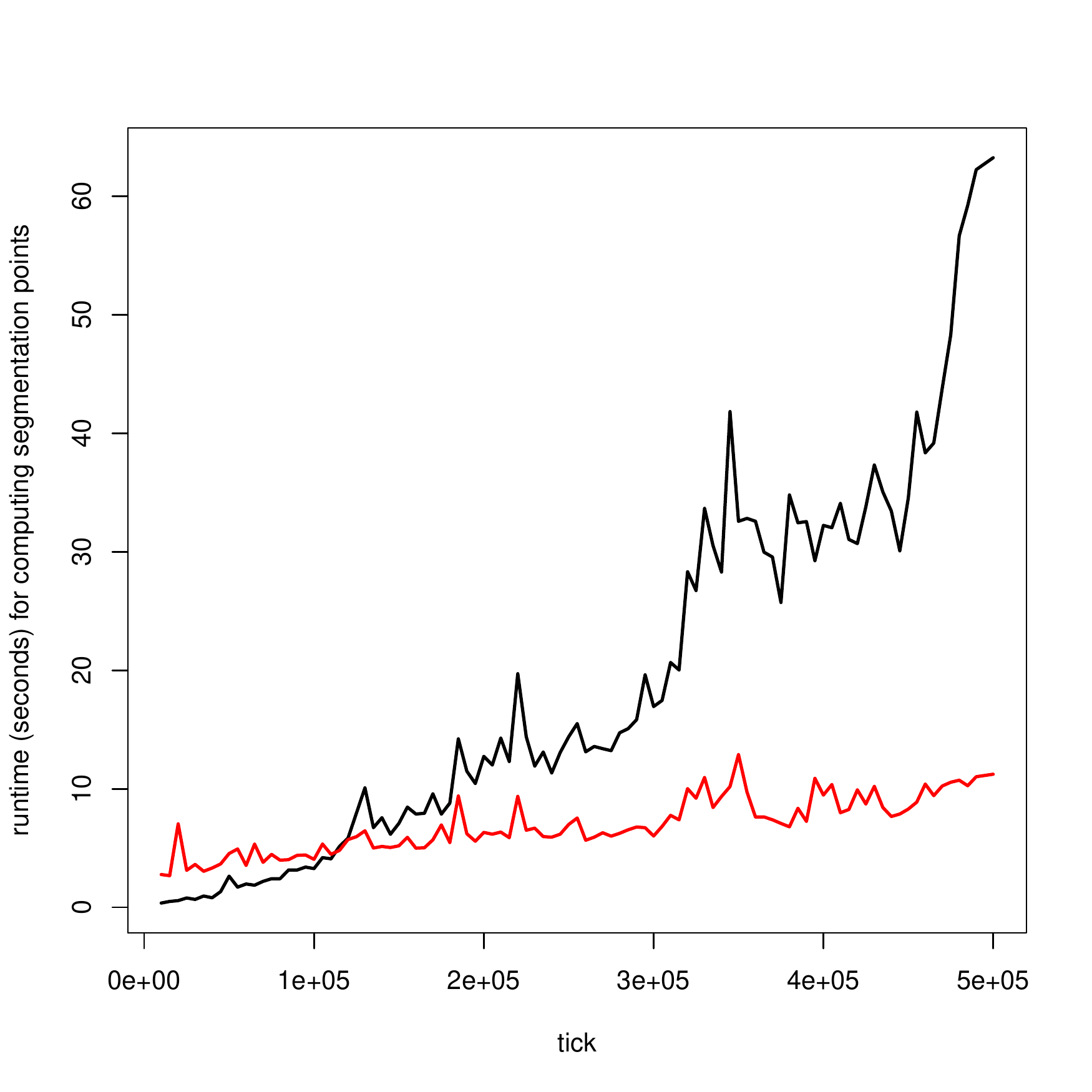}
  \caption{Runtime (seconds) for computing segmentation points via the approximation in Algorithm \ref{alg:query} (red) and via linear programming in Algorithm \ref{alg:general} (black), after every 5000'th element has been added to the time-series in (\ref{equation:sinusoidal}).}\label{fig:runtime_segmentation_approximation}
\endminipage
\end{figure}

\begin{figure}[ht!]
\centering
\includegraphics[width=0.8\linewidth]{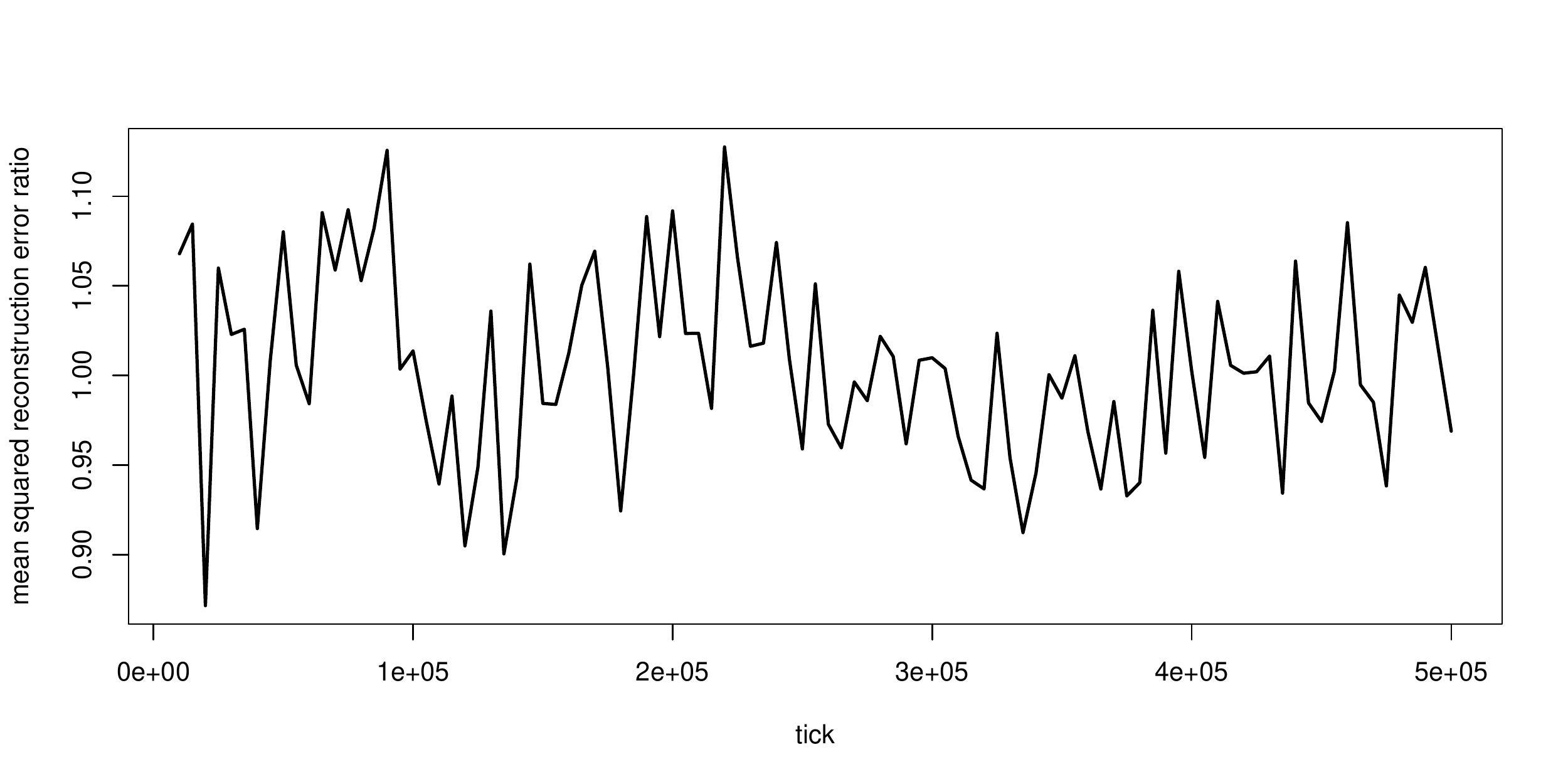}
\caption{The ratio of mean-squared reconstruction error from using piecewise linear reconstructions alongside the segmentation points obtained by both the approximation in Algorithm \ref{alg:query} (red) and via linear programming in Algorithm \ref{alg:general}.}
\label{fig:reconstruction_error_streaming}
\end{figure}

First, Figure \ref{fig:error_over_time_seg_points} shows the relative error of the approximate segmentation points,
$$
\abs*{\tilde{x}_j^S-\tilde{x}_j}/x_{k_{j+1}}, \quad \text{for } j=1,\ldots,n',
$$
with $x_{k_{n'+1}}=n$, after every 5000'th element has been added to the time-series. The theoretical bound in (\ref{equation:errorboundsegpoints}) is also shown. The relative error stays approximately constant over time, and below the bound. Next, Figure \ref{fig:runtime_segmentation_approximation} shows the runtime (in seconds) of computing the segmentation points via the approximation presented in Algorithm \ref{alg:query} within Appendix \ref{sec:appendixstreamingdata}, after every 5000'th element has been added to the time-series. It shows this runtime in comparison to that of computing the actual segmentation points using an implementation of linear programming (Algorithm \ref{alg:general}). Note that the runtime of the approximation is far less than that of re-implementing linear programming each time a new element is added to the time-series for large $n$. This shows the feasibility of applying the segmentation methodology (or an approximation of it) proposed in this paper to time-series obtained from a sensor acquiring data at a fast pace. Finally, Figure \ref{fig:reconstruction_error_streaming} shows the ratio of reconstruction error from using a piecewise linear reconstruction, in (\ref{equation:piecewiselinear}), alongside both the approximate segmentation points and those obtained by continuously implementing Algorithm \ref{alg:general}.  Note that this ratio of errors is approximately equal to one, over the data stream, showing that there is negligible loss in reconstruction accuracy in computing approximate segmentation points instead of using the linear programming algorithm in Algorithm \ref{alg:general}.

\subsection{Accelerometer data}

\begin{figure}[ht!]
\centering
\includegraphics[width=\linewidth]{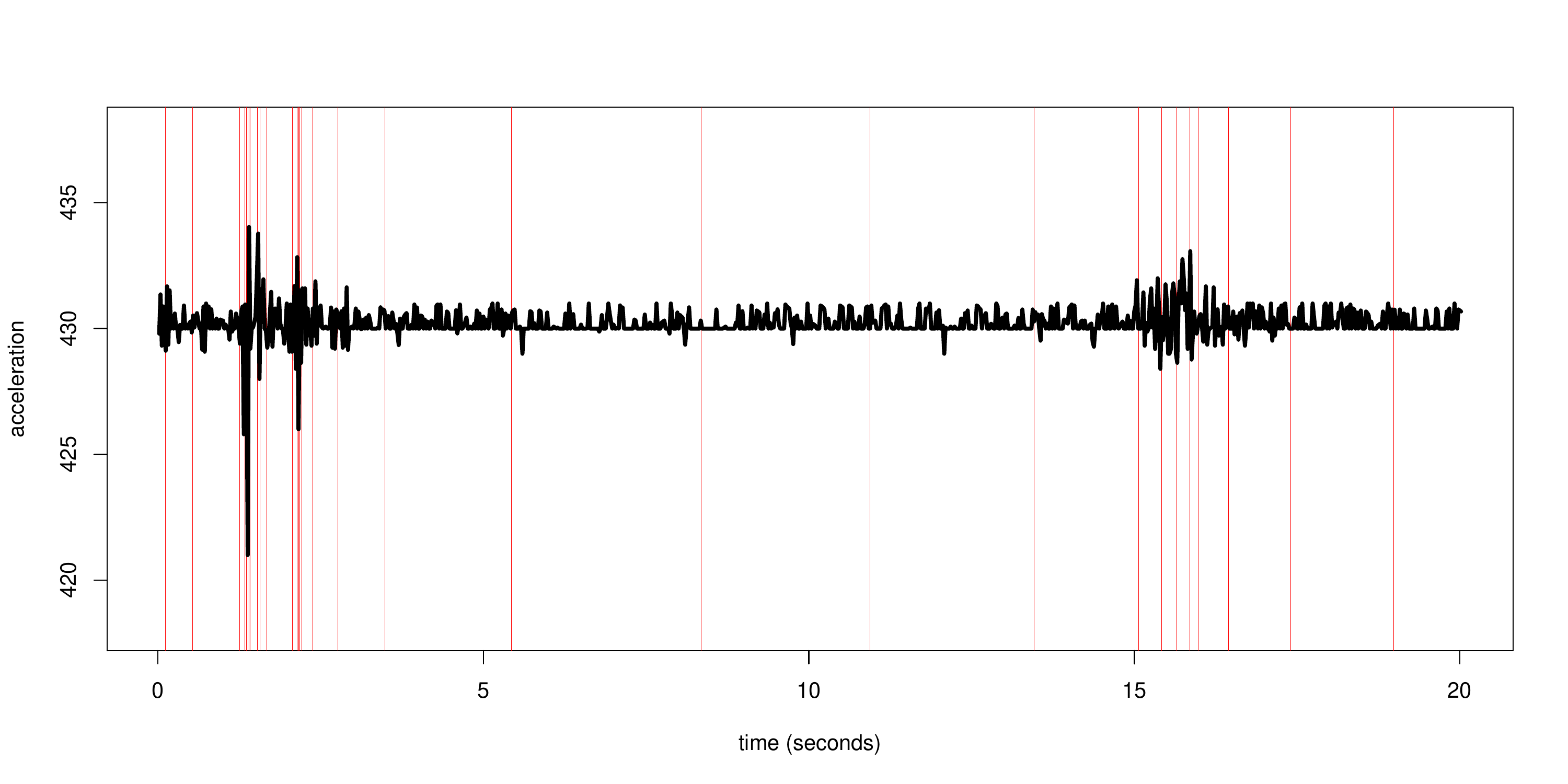}
\caption{A 20 seconds snippet of an accelerometer time-series, where data is acquired at 40Hz. Three signals representing pedestrians interacting with the footbridge are captured in the snippet. Segmentation points, obtained from using the proposed methodology in Sec. \ref{sec:otsegmentation}, are shown by the red vertical lines. Note they cluster around the times at which the three signals occur.}
\label{fig:acceleration_ot}
\end{figure}

This section applies the proposed compression methodology to a time-series $y_1,y_2,\ldots$ generated by accelerometers instrumented on the pedestrian footbridge, introduced in Sec. \ref{sec:data}. Data from one of the sensors (City-side: pi-pier9-bridge-accel-5-9-a-z-9) in the described accelerometer network is considered here. This time-series is acquired at 40Hz over a total time of 20 seconds. There are three signals in the time-series, seemingly corresponding to a pedestrian-event occuring on the bridge near the sensor three times. As aforementioned, accelerometer signals have an oscillatory-like shape, and therefore the relevance score we use to generate the segmentation points in this example corresponds to $\phi_i=\abs{y_i-y_{i-1}}^2$. The intuition behind this choice is that oscillations in the time-series are larger during a signal, rather than during background sensor noise. This can be seen in Figure \ref{fig:acceleration_ot}, where $n'=30$ segmentation points, obtained using the methodology presented in Sec. \ref{sec:otsegmentation}, are also shown.
Notice that the segmentation points gather around the points where the oscillations are largest, that could represent a pedestrian-event being detected by the accelerometer. On the other hand, they are more sparsely spread out at times when there appears to be only sensor noise present. Interestingly, the third and final signal has the least dense concentration of segmentation points out of all three signals, given it does not exhibit as large oscillations as the other two signals do.

\iffalse
\subsubsection{Reconstruction from compression}

\fi

\subsection{Strain sensor data}

\begin{figure}[!htb]
\centering
\minipage{1 \textwidth}
\centering
\includegraphics[width=\linewidth]{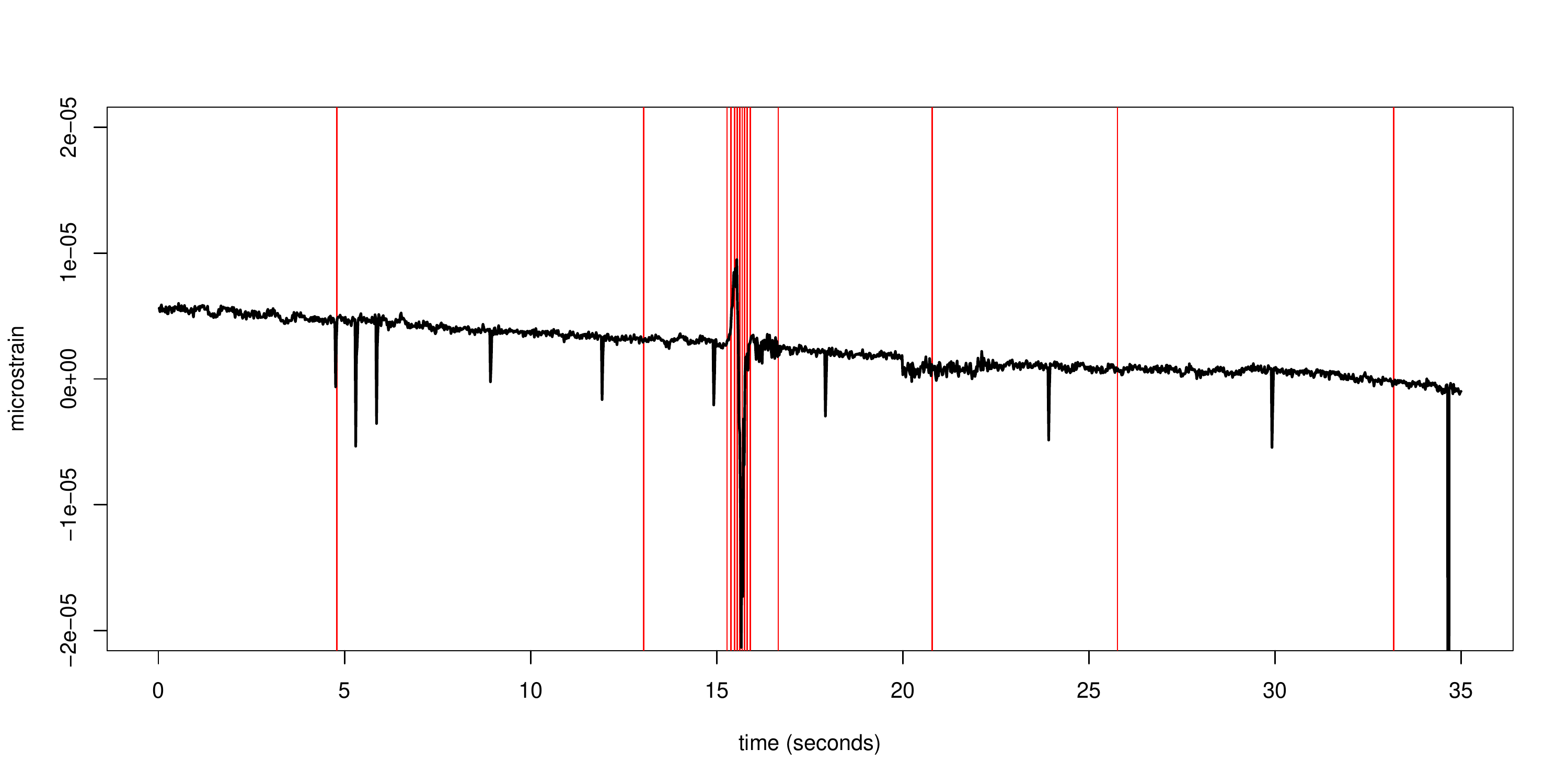}
\caption{A snippet of time-series data obtained from a strain sensor instrumented on the pedestrian bridge described in Sec. \ref{sec:data} (in black). Approximately half way through the snippet of data a pedestrian-event seemingly occurs causing a signal in the time-series. The segmentation points (shown in red), obtained from using the segmentation method proposed in Sec. \ref{sec:otsegmentation} cluster near this signal.}
\label{fig:strain_ot}
\endminipage\vfill
\minipage{1 \textwidth}
\centering
\includegraphics[width=\linewidth]{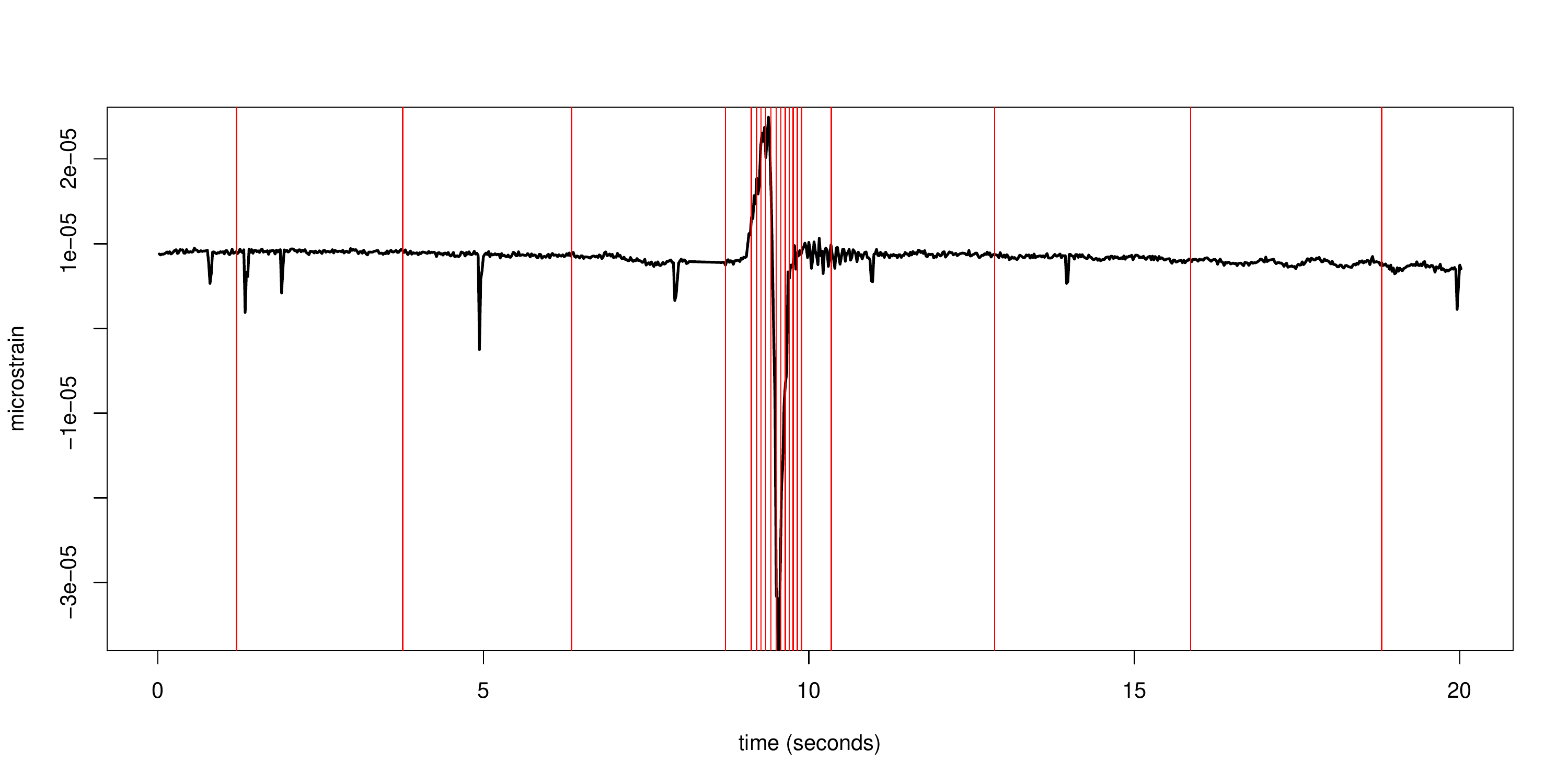}
\caption{The same as Figure \ref{fig:strain_ot}, with a different snippet of time-series data obtained from the strain sensor instrumented on the pedestrian footbridge described in Sec. \ref{sec:data}.}
\label{fig:strain_ot_2}
\endminipage
\end{figure}

This section now applies the proposed compression methodology to a time-series obtained from a strain sensor (City-side / left: pi-pier9-bridge-strain-2-left-s-0) within the network instrumented on the pedestrian bridge introduced in Sec. \ref{sec:data}. Relevant signals within the time-series, seemingly corresponding to a pedestrian-event occuring near the sensor, appear as a sinusoidal-like wave (see Figure \ref{fig:ot_signal_2}). Inspired by the form of this signal, the relevance score used to generate the segmentation points in this example corresponds to that in (\ref{equation:distancetransformfunc}), where the query shape is given by,
$$
q_i=\left(\sigma_{y_j}\sin\left(i\pi/25\right)\right) + \mu_{y_j}, \quad i=1,2,\ldots,51,
$$
where $\mu_{y_j}$ and $\sigma_{y_j}$ are the sample mean and standard deviation of the subsequence $\boldsymbol y_{\max(j-25,1):\min(j+25, n)}$. Therefore normalization is employed here to aid the pattern detection metric in (\ref{equation:distancetransformfunc}). Figures \ref{fig:strain_ot} and \ref{fig:strain_ot_2} show the placements of the segmentation points for two snippets of data from the strain sensor (acquired at 80Hz), each containing a single signal that seemingly corresponds to a pedestrian-event. In both cases, the segmentation points are very sparse for all times that aren't in the immediate interval of the signal; instead they are concentrated on the signal itself.

\begin{figure}[ht!]
\centering
\includegraphics[width=0.8\linewidth]{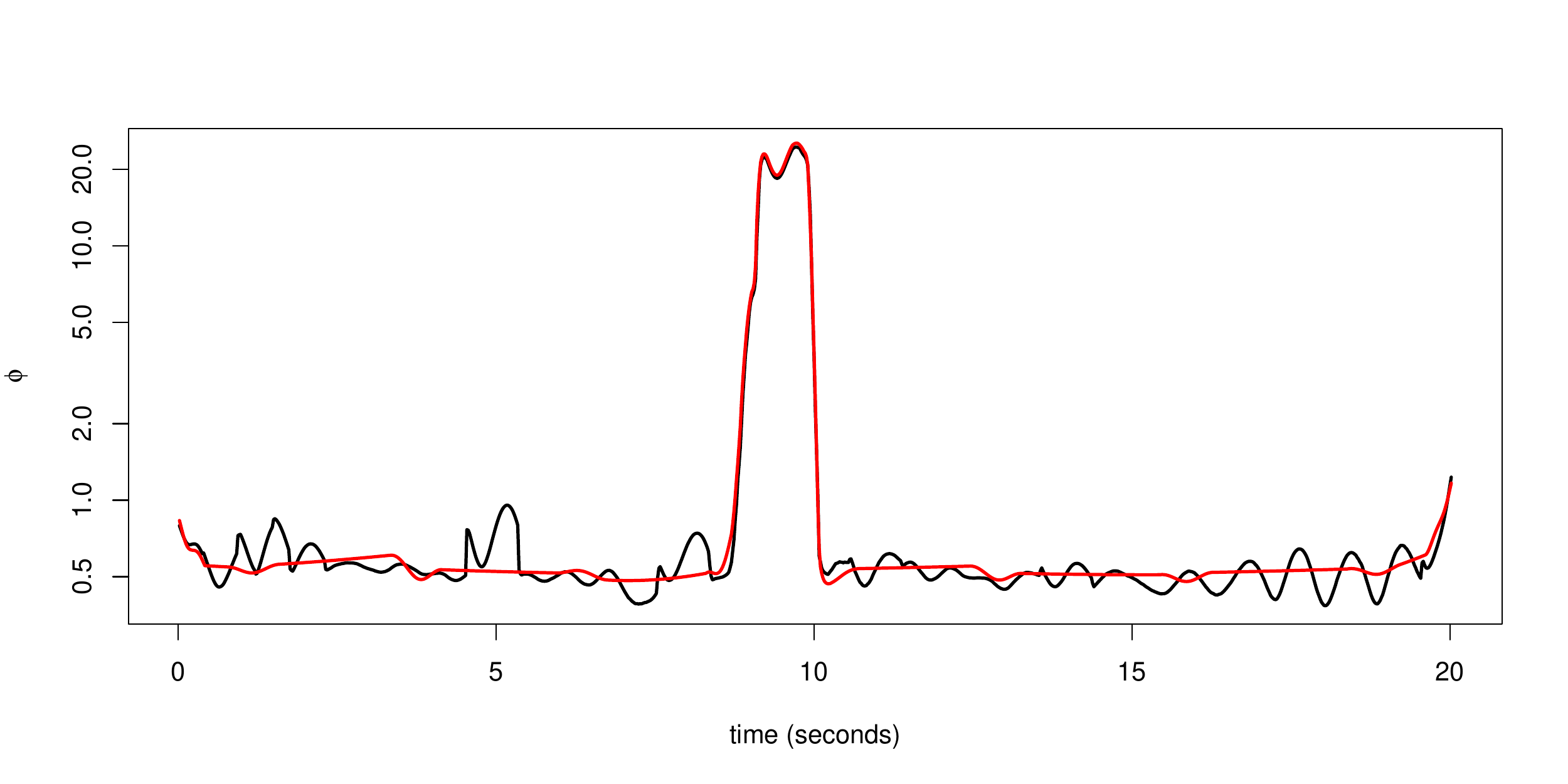}
\caption{The relevance scores $\phi_i$ of the reconstructed piecewise time-series obtained from the segmentation points shown in Figure \ref{fig:strain_ot_2} (in red), alongside the relevance scores of the original time-series (in black).}
\label{fig:reconstruction_ot_z}
\end{figure}

An interesting aspect of the piecewise linear compressed reconstruction from the segmentation points, obtained from using the proposed methodology, is the relevance score for this reconstruction. A piecewise-linear approximation using the segmentation points computed in Figure \ref{fig:strain_ot_2} is obtained, and Figure \ref{fig:reconstruction_ot_z} shows the value of the relevance score in (\ref{equation:distancetransformfunc}) for this approximation alongside that from the original time-series. As one can see, the relevance scores match well for large values (corresponding to $x_i$ values which are close to segmentation points), but do not match well for the lower, less relevant values. This shows the benefit of the proposed methodology at being able to preserve key features of the original time-series, specified by the relevance score used. The error of the relevance score for compressed reconstructions using segmentation points obtained via the proposed methodology is investigated in Sec. \ref{sec:analysisreconstruction}.

\subsubsection{Reconstruction from compression}

\begin{figure}[!htb]
\centering
\minipage{0.48 \textwidth}
\centering
\includegraphics[width=\linewidth]{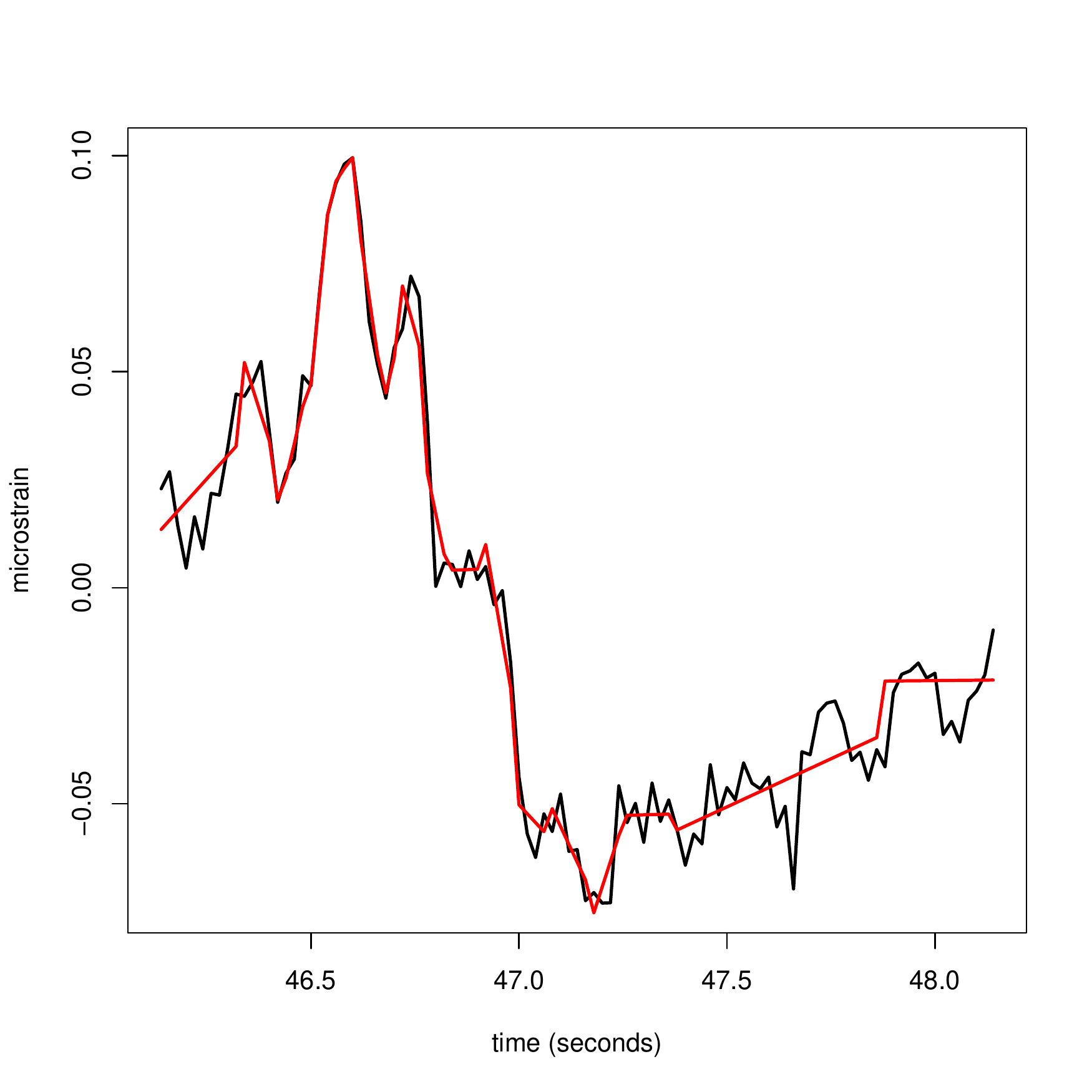}
\caption{A time-series snippet obtained from a strain sensor instrumented on the footbridge described in Sec. \ref{sec:data} for a manually extracted period of time (during which a pedestrian-event occurs). The piecewise reconstruction obtained from using the proposed segmentation methodology in Sec. \ref{sec:otsegmentation} is shown in red.}
\label{fig:event_strain_reconstruction}
\endminipage\hfill
\minipage{0.48 \textwidth}
\centering
\includegraphics[width=\linewidth]{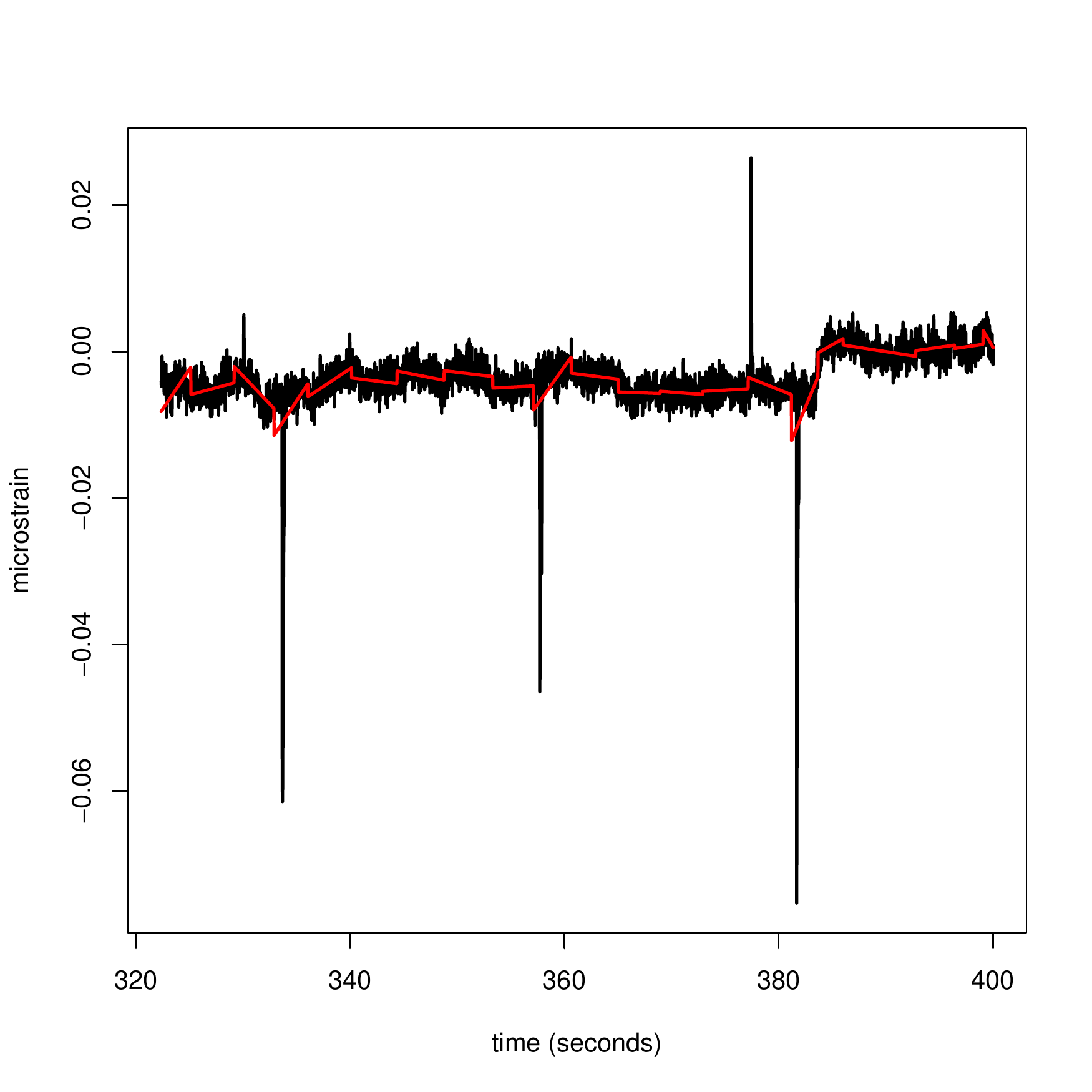}
\caption{A time-series snippet obtained from a strain sensor instrumented on the footbridge described in Sec. \ref{sec:data} for a manually extracted non-event period of time. The piecewise reconstruction obtained from using the proposed segmentation methodology in Sec. \ref{sec:otsegmentation} is shown in red.}
\label{fig:nonevent_strain_reconstruction}
\endminipage
\end{figure}

We will now assess how the compressed reconstruction of a time-series obtained from the strain sensor instrumented to the pedestrian footbridge introduced in Sec. \ref{sec:data}, obtained using the compression methodology in this paper, performs at representing the original time-series in a lower dimensional form. To do this, we will concentrate on assessing the reconstruction error and space-efficiency within parts of the original strain sensor time-series that are highly relevant to the analyzer: the signals corresponding to pedestrian-events. A 400 second long time-series is obtained from the strain sensor. This snippet contained 5 pedestrian-event signals. Five 2 second long intervals, containing these signals, were extracted manually, and the non-event periods in between these intervals were recorded separately. Segmentation points were computed as in the previous section, and a compressed reconstruction was obtained using a piecewise regression approximation to the original time-series. The reconstruction during one of the extracted event intervals containing a signal is shown in Figure \ref{fig:event_strain_reconstruction}, in addition to the reconstruction during one of the non-event periods in between the extracted intervals shown in Figure \ref{fig:nonevent_strain_reconstruction}. Compression ratios, and the average relative squared reconstruction error, $\frac{1}{|J|}\left(\sum_{t\in J}\abs{S(x_t;\big\{\tilde{x}_j\big\}_{j=1}^{n'})-y_t}^2\right)/\left(\sum_{t \in J}\abs{y_t}\right)$, where $J$ corresponds to all indices within a particular time interval of the reconstruction, were computed for each extracted event interval and each non-event period in between the extracted intervals. These are shown in Table \ref{tab:reconstructions}. One can note from the aforementioned figures and this table that the reconstruction is refined, leading to lower error, in the periods of high relevance (signals seemingly caused by pedestrian events). Compression ratios are lower in these periods, than during the non-event periods, as more segmentation points have concentrated on them. The higher compression ratios in the non-event periods, coupled with the lower error during the extracted event intervals, show the effectiveness of the reconstruction at producing very similar values to the original time-series during relevant periods only in a much lower dimensional form.

\begin{table}[ht!]
\centering
  \begin{tabular}{ | c | c | c |}
    \hline
    Period & Compression Ratio & Relative Squared Reconstruction Error \\ \hline \hline
    Non-event 1 &183.1&0.99 \\ \hline
	Non-event 2 &168.5&0.24 \\ \hline
	Non-event 3 &143.4&0.41 \\ \hline
	Non-event 4 &149.0&0.44 \\ \hline
	Non-event 5 &190.5&0.41 \\ \hline
	Event 1 &5.31&0.02 \\ \hline
	Event 2 &5.94&0.03 \\ \hline
	Event 3 &4.80&0.01 \\ \hline
	Event 4 &4.81&0.02 \\ \hline
	Event 5 &9.18&0.05 \\ \hline
	\hline
  \end{tabular}
  \caption{Compression ratios and relative squared reconstruction errors during five non-event periods and five pedestrian-event periods for a time-series obtained from the strain sensor on the pedestrian footbridge described in Sec. \ref{sec:data}. Period start and end points were chosen manually.}
  \label{tab:reconstructions}
  \end{table}

%%% Local Variables: 
%%% mode: latex
%%% TeX-master: "optimal_transport_compression"
%%% End:

%% file: conclusion.tex
% summation
This paper has presented a compression technique for data streamed from instrumented infrastructure, such as bridges, roads and tunnels fitted with sensor networks, for applications including condition and structural health monitoring. Especially when data is acquired frequently, relative to any changes exhibited in the structure, it is important to wisely compress data down to a manageable quantity for storage and analysis. Methodology is presented for cases where data is given in a single batch, and where data is acquired sequentially in an indefinite stream. The proposed compression technique produces a piecewise aggregate approximation (segmented time-series) that preserves user-defined particular patterns or features that exist in the original time-series. This paper uses the motivating example of particular patterns of signals from accelerometers and strain sensors instrumented on a pedestrian footbridge, that could represent a pedestrian-event (such as a person walking) in the vicinity of these sensors, as the features that one would like to preserve in a compression.

% more about methodology
The methodology works as follows. A user-defined relevance score is first used to create weights for each data point in a time-series; points are weighted relative to how important it is to the appearance of features or patterns that one would like to preserve in the compression. Then optimal transport is used to find the optimal piecewise segmentation that preserves sequences of points within the time-series that have high relevance. This can be done via linear programming which can be implemented quickly even for relatively large data-sets. In the case where the data is streamed sequentially over time (e.g. from a sensor network instrumented on an operational structure), a bounded approximation to the optimal piecewise segmentation can be maintained over time and queried in a significantly reduced runtime relative to re-computing the linear programming result.

% problems and future research metrics
The features that the compression should preserve inform the choice of the relevance score used alongside the proposed methodology. For example, similarity search and distance measures can be used to preserve a particular query shape or pattern in the time-series data. Future extensions of this work should explore the properties of compressions constructed using the proposed methodology alongside more exoctic relevance scores (e.g. Markov-models \citep{Ge} or probabilistic warping \citep{Bautista}) and compositions of relevance scores. It only seems natural that by doing the latter for example, one could extend this methodology to compressing a time-series whilst preserving multiple important features or patterns, such as data acquired from sensors that produce different signals for different types of events (e.g. railway bridges, where different types of trains frequently pass over it).

% how important to field, and key numerical demonstration results to show impact in field
The motivating example of applying the proposed methodology to compressing data-sets obtained from a pedestrian footbridge instrumented with strain sensors and accelerometers is considered via a series of numerical demonstrations towards the end of this paper. These demonstrations highlight the effectiveness of the compression at preserving key features (e.g. a sinusoidal-type wave of measurements) in the original data from the strain sensors and accelerometers that represent a pedestrian-event near the sensor location. Due to the choice of these relevance scores for the implementation of the proposed methodology, the compression ignores large-magnitude noise and outliers (possibly due to electrical currents) that often make traditional analyzes of raw data obtained from strain sensors and accelerometers difficult.  The reconstructed compressed data obtained from this methodology exhibits low error, with respect to the original data, during occurences of pedestrian-events, and high compression ratios (a metric for the space-efficiency of the compression) during unimportant periods of data. These demonstrated properties are necessary in alleviating the high complexity of storing and analyzing streaming sensor data from instrumented infrastructure. Therefore this work contributes towards important research efforts to improve structural health and condition monitoring systems used alongside novel and contemporary sensing technologies.

%%% Local Variables: 
%%% mode: latex
%%% TeX-master: "optimal_transport_compression"
%%% End: